\documentclass{article} 
\usepackage{iclr2025_conference,times}

\usepackage{wrapfig}
\usepackage{lipsum} 
\usepackage{wrapfig}

\usepackage{booktabs}      
\usepackage{multirow}      
\usepackage{amssymb}       
\usepackage{graphicx, subfigure}     

\usepackage{colortbl}
\usepackage{xcolor}

\usepackage{hyperref}

\usepackage{url}
\usepackage{multicol}
\usepackage{bm}
\usepackage{aliascnt}
\usepackage{adjustbox}
\usepackage{siunitx} 
\usepackage{booktabs}
\usepackage{multirow} 
\usepackage{enumitem}
\usepackage{booktabs} 
\usepackage{amssymb}  
\usepackage{amsmath}  
\usepackage{graphicx} 
\usepackage{longtable} 
\usepackage{graphicx}
\usepackage{subcaption}
\usepackage{calc}
\usepackage[most]{tcolorbox}
\usepackage{listings}
\usepackage{wrapfig}
\usepackage{caption}

\usepackage[dvipsnames]{xcolor}  
\definecolor{highlightgray}{gray}{0.92} 
\definecolor{myblue}{RGB}{235, 235, 250}
\definecolor{mygreen}{rgb}{0.92, 1.0, 0.92}
\definecolor{top1}{HTML}{F7CDC8}
\definecolor{top2}{HTML}{D1EDF9}
\definecolor{top3}{HTML}{FDF1D0}
\definecolor{uclablue}{rgb}{0.15, 0.45, 0.68}
\hypersetup{
    breaklinks,
    citecolor=uclablue,
    colorlinks=true,
}

\newtcolorbox{AIbox}[2][]{aibox,title=#2,#1}
\tcbset{
  aibox/.style={
    top=10pt,
    colframe=black,
    colbacktitle=black,
    coltitle=white,
    center,
  }
}

\lstdefinelanguage{prompt}{
    basicstyle=\scriptsize\ttfamily, 
    mathescape=true,        
    escapebegin=\color{latentcolor},  
    escapeend={},
    escapechar=@,
    stringstyle = \color{myorange},
    showstringspaces = false,
    moredelim = [s][\color{mypink}]{`}{`},
    moredelim = [s][\color{mybrown}]{```json}{```},
    moredelim = [s][\color{latentcolor}]{<StartOfLatent>}{<EndOfLatent>},
    literate = %
        {\ \ a.\ }{{\textcolor{mypurple}{\ \ a.\ }}}5
        {\ \ b.\ }{{\textcolor{mypurple}{\ \ b.\ }}}5
        {\ \ c.\ }{{\textcolor{mypurple}{\ \ c.\ }}}5
        {\ \ d.\ }{{\textcolor{mypurple}{\ \ d.\ }}}5
        {\ \ e.\ }{{\textcolor{mypurple}{\ \ e.\ }}}5
        {\ \ f.\ }{{\textcolor{mypurple}{\ \ f.\ }}}5
        {\ \ g.\ }{{\textcolor{mypurple}{\ \ g.\ }}}5
        {\ \ h.\ }{{\textcolor{mypurple}{\ \ h.\ }}}5
        {\ I.\ }{{\textcolor{mypurple}{\ I.\ }}}4
        {\ II.\ }{{\textcolor{mypurple}{\ II.\ }}}5
        {\ III.\ }{{\textcolor{mypurple}{\ III.\ }}}6
        {\ IV.\ }{{\textcolor{mypurple}{\ IV.\ }}}5
        {\ V.\ }{{\textcolor{mypurple}{\ V.\ }}}4
}

\newtcblisting[auto counter, number within=subsection]{prompt}[4][]{%
  before skip=6pt, after skip=6pt,    
  top=6pt, bottom=6pt, left=7pt, right=7pt, 
  enhanced, breakable,
  colback=white, colframe=gray!65,
  boxrule=0.6pt, arc=2pt,
  drop shadow={black!15},             
  listing only,
  listing options={
    language=prompt,
    upquote=true,
    basicstyle=\scriptsize\ttfamily \setlength{\baselineskip}{1.1\baselineskip},
    columns=fullflexible,
    breaklines=true,
    breakindent=0pt,
    xleftmargin=\ifx\empty#2\empty0pt\else#2\fi,
    xrightmargin=\ifx\empty#3\empty0pt\else#3\fi,
    aboveskip=0pt,                    
    belowskip=0pt,                    
    showstringspaces=false,
  },
  title={\ifstrempty{#4}{Prompt \thetcbcounter}{Prompt \thetcbcounter: #4}},
  colbacktitle=gray!6, coltitle=black,
  attach boxed title to top left={yshift=-1mm, xshift=6pt},
  boxed title style={colback=gray!6, boxrule=0pt, sharp corners},
  #1
}

\newtcblisting[auto counter, number within=subsection]{showcase}[4][]{%
  before=\par\vspace{\baselineskip},
  after=\par,
  width=\linewidth,
  enhanced,
  arc=0em,
  boxrule=1pt,
  listing only,
  listing options={
    language=prompt,
    upquote=true,
    basicstyle=\scriptsize\ttfamily \setlength{\baselineskip}{1.1\baselineskip},
    breaklines=true,
    breakindent=0pt,
    xleftmargin=\ifx\empty#2\empty-12pt\else#2\fi,
    xrightmargin=\ifx\empty#3\empty-5pt\else#3\fi,
    aboveskip=-4pt,
    belowskip=-4pt,
    columns=fullflexible,
    },
  colback=white,
  colframe=gray,
  colbacktitle=gray!5,
  coltitle=black,
  attach boxed title to top center={yshift=-3mm},
  box align=center,
  parbox=false,
  title={\ifstrempty{#4}{Example \thetcbcounter}{Example \thetcbcounter: #4}},
  #1
}

\definecolor{linkColor}{rgb}{0.2,0.4,0.6}
\definecolor{myblue}{HTML}{0379AC}
\definecolor{myred}{HTML}{A50E50}
\definecolor{myorange}{RGB}{238, 133, 74}
\definecolor{latentcolor}{named}{cyan}
\definecolor{normalcolor}{RGB}{0, 0, 0}
\usepackage{marvosym}
\definecolor{lightblue1}{rgb}{0.97, 0.985, 1} 
\definecolor{lightblue2}{rgb}{0.92, 0.965, 1} 
\definecolor{lightblue3}{rgb}{0.84, 0.93, 1}
\definecolor{lightblue4}{rgb}{0.74, 0.87, 1}
\definecolor{lightblue5}{rgb}{0.64, 0.81, 1}
\definecolor{lightblue6}{rgb}{0.54, 0.75, 1}

\definecolor{lightgreen1}{rgb}{0.97, 1.00, 0.97}
\definecolor{lightgreen2}{rgb}{0.92, 0.98, 0.92}
\definecolor{lightgreen3}{rgb}{0.84, 0.95, 0.84}
\definecolor{lightgreen4}{rgb}{0.74, 0.91, 0.74}
\definecolor{lightgreen5}{rgb}{0.64, 0.86, 0.64}
\definecolor{lightgreen6}{rgb}{0.54, 0.81, 0.54}

\definecolor{lightorange1}{rgb}{1.00, 0.98, 0.95}
\definecolor{lightorange2}{rgb}{1.00, 0.95, 0.85}
\definecolor{lightorange3}{rgb}{1.00, 0.90, 0.70}
\definecolor{lightorange4}{rgb}{1.00, 0.85, 0.55}
\definecolor{lightorange5}{rgb}{1.00, 0.80, 0.40}
\definecolor{lightorange6}{rgb}{1.00, 0.75, 0.30}

\definecolor{lightpurple1}{rgb}{0.985, 0.97, 1.00}
\definecolor{lightpurple2}{rgb}{0.96, 0.92, 1.00}
\definecolor{lightpurple3}{rgb}{0.93, 0.84, 1.00}
\definecolor{lightpurple4}{rgb}{0.87, 0.74, 1.00}
\definecolor{lightpurple5}{rgb}{0.81, 0.64, 1.00}
\definecolor{lightpurple6}{rgb}{0.75, 0.54, 1.00}

\definecolor{lightred1}{rgb}{1.00, 0.97, 0.97}
\definecolor{lightred2}{rgb}{1.00, 0.92, 0.92}
\definecolor{lightred3}{rgb}{1.00, 0.84, 0.84}
\definecolor{lightred4}{rgb}{1.00, 0.74, 0.74}
\definecolor{lightred5}{rgb}{1.00, 0.64, 0.64}
\definecolor{lightred6}{rgb}{1.00, 0.54, 0.54}

\definecolor{lightcyan1}{rgb}{0.97, 1.00, 1.00}
\definecolor{lightcyan2}{rgb}{0.92, 0.98, 0.98}
\definecolor{lightcyan3}{rgb}{0.84, 0.95, 0.96}
\definecolor{lightcyan4}{rgb}{0.74, 0.91, 0.94}
\definecolor{lightcyan5}{rgb}{0.64, 0.87, 0.92}
\definecolor{lightcyan6}{rgb}{0.54, 0.83, 0.90}

\usepackage{xcolor,colortbl}
\definecolor{Gray}{gray}{0.85}
\definecolor{LightCyan}{rgb}{0.88,1,1}
\definecolor{greyC}{RGB}{180,180,180}
\definecolor{greyL}{RGB}{235,235,235}
\definecolor{citeColor}{RGB}{0,20,115}
\hypersetup{colorlinks,linkcolor={red},citecolor={citeColor},urlcolor={citeColor}}
\definecolor{shadecolor}{rgb}{0.92,0.92,0.92}

\usepackage{graphicx}
\usepackage{subcaption}
\usepackage{url}
\newcommand{\method}{UniCom}

\usepackage{booktabs}
\usepackage{multirow}
\usepackage{cleveref}
\crefname{template}{Template}{Template}

\usepackage{enumitem} 
\usepackage[most]{tcolorbox}
\usepackage{pifont}
\definecolor{rliableblue}{RGB}{0, 102, 204} 

\tcbset{
    myblueboxstyle/.style={
        colback=lightblue3, 
        colframe=lightblue5, 
        coltitle=black, 
        fonttitle=\bfseries, 
        boxrule=1pt, 
        width=\linewidth, 
        left=2pt, 
        right=2pt, 
        top=2pt, 
        bottom=2pt, 
        sharp corners, 
        boxsep=2pt
    }
}

\lstdefinestyle{iclrstyle}{
    language=Python,
    basicstyle=\ttfamily\small,  
    columns=fullflexible,        
    keepspaces=true,             
    showspaces=false,            
    showstringspaces=false,      
    commentstyle=\color{gray}\itshape, 
    keywordstyle=\color{codekw}\bfseries, 
    stringstyle=\color{myorange}, 
    escapechar=|,                
    frame=none,                  
    xleftmargin=1.5em,           
    aboveskip=0.5em,             
    belowskip=0.5em,             
    breaklines=true, 
    breakindent=0pt,
}

\usepackage{algorithm}
\usepackage{algpseudocode}
\usepackage{listings} 
\usepackage{xcolor}
\usepackage{etoolbox}
\makeatletter
\AfterEndEnvironment{algorithm}{\let\@algcomment\relax}
\AtEndEnvironment{algorithm}{\kern2pt\hrule\relax\vskip3pt\@algcomment}
\let\@algcomment\relax
\newcommand\algcomment[1]{\def\@algcomment{\footnotesize#1}}
\renewcommand\fs@ruled{\def\@fs@cfont{\bfseries}\let\@fs@capt\floatc@ruled
  \def\@fs@pre{\hrule height.8pt depth0pt \kern2pt}%
  \def\@fs@post{}%
  \def\@fs@mid{\kern2pt\hrule\kern2pt}%
  \let\@fs@iftopcapt\iftrue}
\makeatother

\NewDocumentCommand{\xx}
{ mO{} }{\textcolor{blue}{\textsuperscript{\textit{todo}}\textsf{\textbf{\small[#1]}}}}
\usepackage{xspace}

\usepackage{soul}
\definecolor{codeblue}{rgb}{0.25,0.5,0.5}
\definecolor{codekw}{rgb}{0.85, 0.18, 0.50}
\definecolor{diffgreen}{rgb}{0.0, 0.6, 0.0} 
\definecolor{diffred}{rgb}{0.8, 0.0, 0.0}

\title{{\method}: Unified Multimodal Modeling via \\
  Compressed Continuous Semantic Representations}

\author{
Yaqi Zhao$^{1,3*}$, 
Wang Lin$^{2,3*}$, 
Zijian Zhang$^{3}$, 
Miles Yang$^{3}$, \\
Jingyuan Chen$^{2,\dagger}$, 
Wentao Zhang$^{1,\dagger}$,
Zhao Zhong$^{3}$,
Liefeng Bo$^{3}$\\
\textbf{$^1$Peking University} \quad \textbf{$^2$Zhejiang University} \quad \textbf{$^3$Tencent Hunyuan}\\[0.7em]
Project Page: \href{https://miazhao7708.github.io/UniComPage/}{\texttt{https://miazhao7708.github.io/UniComPage/}}}

\begin{document}
\maketitle
\renewcommand*{\thefootnote}{\fnsymbol{footnote}}
\footnotetext{* Equal contribution. Work done during internship at Tencent Hunyuan.}
\footnotetext{$\dagger$ Corresponding Authors.}

\begin{abstract}
Current unified multimodal models typically rely on discrete visual tokenizers to bridge the modality gap. However, discretization inevitably discards fine-grained semantic information, leading to suboptimal performance in visual understanding tasks. Conversely, directly modeling continuous semantic representations (e.g., CLIP, SigLIP) poses significant challenges in high-dimensional generative modeling, resulting in slow convergence and training instability. To resolve this dilemma, we introduce UniCom, a unified framework that harmonizes multimodal understanding and generation via compressed continuous representation. 
We empirically demonstrate that reducing channel dimension is significantly more effective than spatial downsampling for both reconstruction and generation. Accordingly, we design an attention-based semantic compressor to distill dense features into a compact unified representation. 
Furthermore, we validate that the transfusion architecture surpasses query-based designs in convergence and consistency. 
Experiments demonstrate that UniCom achieves state-of-the-art generation performance among unified models. Notably, by preserving rich semantic priors, it delivers exceptional controllability in image editing and maintains image consistency even without relying on VAE. 
\end{abstract}

\section{Introduction}
The pursuit of unified multimodal models~\cite{pan2025generative,deng2025emerging,tong2026scaling}, capable of both understanding and generating content across modalities such as text and images, represents a significant paradigm shift in artificial intelligence. A key insight from this evolution is that unification is not merely an engineering challenge of connecting modules. It fundamentally questions the representational divide between discrete, symbolic reasoning (language) and continuous, perceptual synthesis (vision). The ongoing research is, in essence, searching for a ``unified token''—a common representational substrate that can seamlessly flow between understanding and generation, between text and pixels, while being efficiently processed by a scalable neural architecture.

Achieving such unification critically depends on how visual information is represented, making the design of visual encoders a central challenge. 
Some compromise approaches~\cite{chen2025janus,wang2025skywork,qu2025tokenflow,xie2025show,deng2025emerging} adopt hybrid encoders that combine variational autoencoder (VAE~\cite{rombach2022high}) latents with Visual Transformer (ViT) features. While this design facilitates generation, it introduces an inherent representational divergence: semantic understanding and image synthesis are grounded in different feature spaces, fundamentally limiting deeper unification. 
To address this representational schism, some works~\cite{geng2025x,ma2025unitok,han2025vision} move toward using only ViT-based visual embeddings, aiming to anchor both understanding and generation within a single semantic space. Within this line of research, some approaches discretize powerful continuous features (\textit{e.g.}, CLIP/SigLIP~\cite{radford2021learning,tschannen2025siglip2multilingualvisionlanguage}) via vector quantization to obtain tractable visual tokens. Although discretization simplifies generative modeling, it inevitably incurs irreversible information loss, particularly in fine-grained spatial and textural details that are essential for high-fidelity image synthesis.

More recent methods~\cite{zheng2025diffusion,tong2026scaling,gao2025one,chen2025vugen}, abandon quantization altogether and instead operate directly on continuous ViT representations. However, this shift exposes a new challenge: the resulting high-dimensional continuous feature manifold is complex, often non-smooth, and difficult to model generatively (see Fig~\ref{fig:training_dynamics}). Therefore, we revisit the visual embeddings from the two perspectives: \textbf{modeling} and \textbf{prediction}.

For visual embeddings modeling, we introduce a continuous semantic compressor, a lightweight module that learns to project high-dimensional visual semantics into a compact and continuous latent space. This space serves as a unified representation that preserves essential semantic, structural, and textural information while significantly simplifying the underlying data distribution. We systematically compare different projection strategies and evaluate their reconstruction fidelity, demonstrating that visual embeddings can be compressed in a near-lossless manner. Our preliminary experiments suggest that such continuous compression preserves both understanding (see Tab~\ref{tab:vqa_bench_ablation}) and pixel-level reconstruction (see Tab~\ref{tab:image-recon}) capabilities, while substantially improving training stability and computational efficiency.

Given the compressed representation, we next investigate how to effectively predict visual embeddings from textual conditions. Within a single unified framework, we study and compare two distinct generative pathways. One inspired by Transfusion~\cite{zhou2024transfusion} that is trained end-to-end using a unified flow-matching objective over mixed-modality sequences, directly predicting compressed visual latents. Another way introduces a query-based~\cite{pan2025transfer,chen2025blip3} approach, in which a set of learnable queries is used to extract conditional signals 
from a powerful multimodal large language model, thereby leveraging its semantic understanding and reasoning capabilities to guide visual generation. Our comparative analysis reveals that the Transfusion pathway achieves faster convergence and maintains stronger consistency in editing tasks. Therefore, we adopt it as the prediction mechanism of our final model.

Finally, we propose \textbf{UniCom}, a unified large-scale multimodal model that performs generation directly over compressed visual embeddings. Through systematic analysis, we identify that compressing high-dimensional features along the channel dimension, rather than reducing the token sequence, optimally preserves both semantic fidelity and pixel-level details. The resulting compact representation enables highly efficient and stable training, while our unified prediction pathway achieves state-of-the-art performance across diverse tasks. Notably, UniCom excels in fine-grained text rendering and complex image editing without relying on VAE features for identity preservation, demonstrating that semantically rich and well-compressed visual embeddings can serve as an effective universal interface for both understanding and generation.
Our contributions are as follows:
\vspace{-0.5em}
\begin{itemize}[leftmargin=*, itemsep=0.3em, parsep=0.2em]
    \item 
    We establish an effective paradigm for unifying visual understanding and generation by learning to predict continuous, compressed semantic embeddings, which we demonstrate preserves both high-level semantics and fine-grained visual details better than prior quantization methods.
    \item 
    We revealing that compression along the channel dimension is significantly superior to sequence reduction for preserving information, and that an attention-based projector is crucial for maintaining semantic structure.
    \item 
    The proposed UniCom model validates this paradigm at scale, achieving state-of-the-art or competitive performance across image reconstruction, text-to-image generation, and challenging image editing tasks.
\end{itemize}

\begin{figure*}
    \centering
    \includegraphics[width=\linewidth]{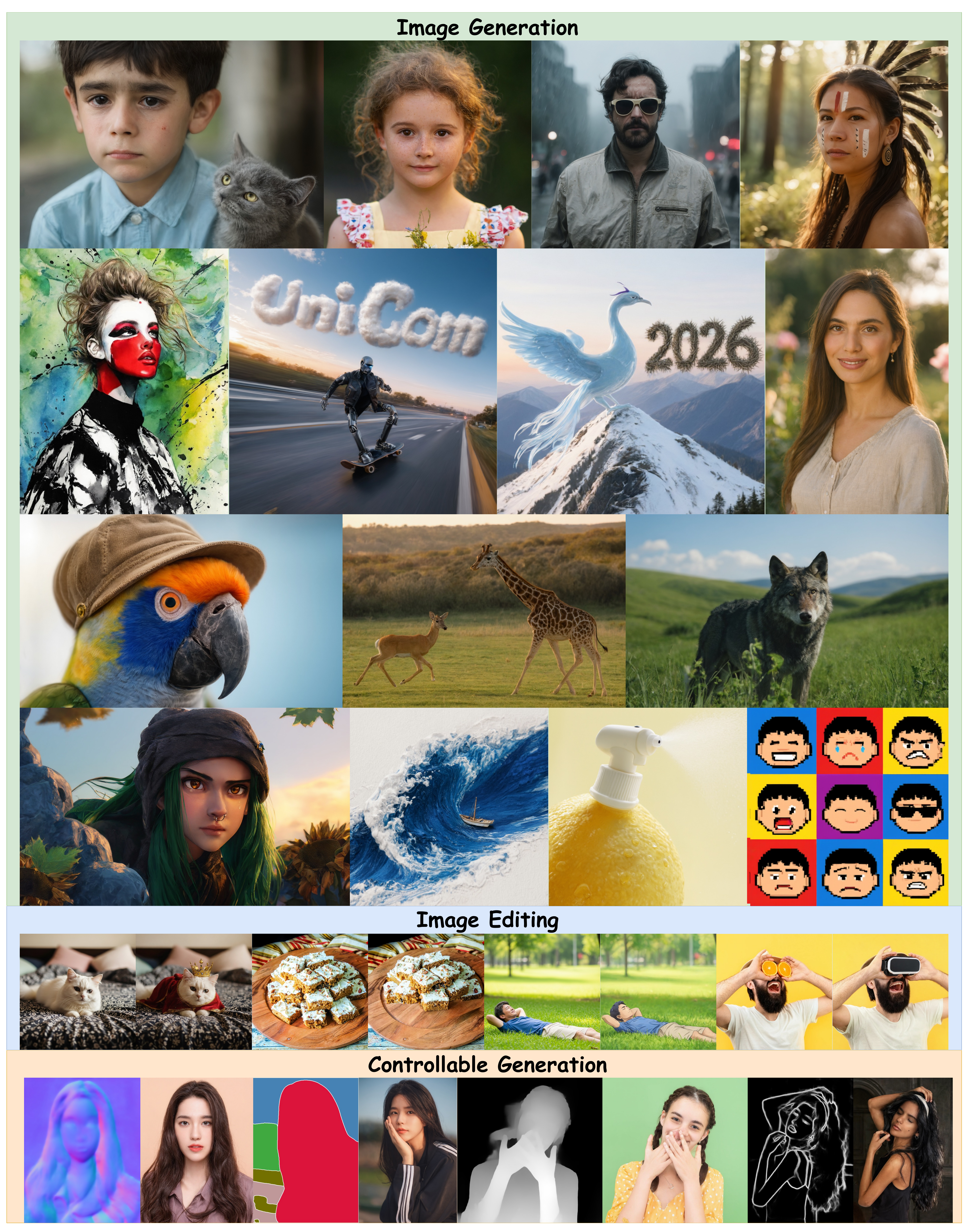}
    \caption{High-quality samples generated by UniCom. Built on compressed continuous representations, our unified multimodal model demonstrates exceptional capabilities in text-to-image generation, precise image editing, and fine-grained controllable generation.}
    \label{fig:case_generation}
\end{figure*}
\section{Related Work}
A prominent approach attempts~\cite{team2024chameleon,wang2024emu3,pan2025generative} to integrate image generation into the autoregressive framework of LLMs, addressing the challenge of image tokenization. Early methods~\cite{geng2025x,gupta2022metamorph,ma2025unitok,chen2025unicode}, serialized vision-language encoders like CLIP/SigLIP into token sequences similar to text, typically paired with a diffusion decoder to combine AR understanding with diffusion-based generation. However, the resulting quantization loss and long sequences often lacked the semantic alignment necessary for efficient cross-modal reasoning. 
Some approaches~\cite{ai2025ming,tang2025unilip,chen2025blip3}, like MetaQuery~\cite{pan2025transfer}, generate query tokens via an LLM, which are then reconstructed to compact visual representations for image decoding through a diffusion model.
Another line of work~\cite{shi2024lmfusion,xie2025show,ma2025janusflow,tong2026scaling}, such as Transfusion~\cite{zhou2024transfusion}, combines AR's sequential reasoning with diffusion's high-fidelity image synthesis. In these models, text tokens are generated autoregressively by an LLM, while image tokens undergo denoising conditioned on the textual context. 
\section{UniCom}
\subsection{Problem Formulation and Unified Framework}
Our work posits that the key to a unified and efficient model lies not in circumventing the VLM's semantic space, but in \emph{transforming it} into a form amenable to generative modeling while preserving its rich informational content. We formulate a two-stage generative process that decomposes the conditional image distribution $P(\mathbf{x} \mid \mathbf{c})$ (where $\mathbf{c}$ is a conditioning context such as text) as follows:
\begin{equation}
P(\mathbf{x} \mid \mathbf{c}) = \int P(\tilde{\mathbf{z}} \mid \mathbf{c}) \cdot P(\mathbf{x} \mid \tilde{\mathbf{z}})\, d\tilde{\mathbf{z}} .
\end{equation}

Here, $\tilde{\mathbf{z}} \in \mathbb{R}^{N \times d}$ (with $d \ll D$) represents the image in a compressed semantic latent space $\tilde{\mathcal{Z}}$. This formulation introduces a critical intermediate variable $\tilde{\mathbf{z}}$ with two properties:
\begin{enumerate}
    \item \textbf{Semantic Fidelity:} $\tilde{\mathbf{z}}$ retains the necessary information from $\mathbf{Z}$ to reconstruct the image $\mathbf{x}$ with high fidelity, ensuring a strong connection to understanding.
    \item \textbf{Generative Tractability:} The distribution $P(\tilde{\mathbf{z}} \mid \mathbf{c})$ over the lower-dimensional space $\tilde{\mathcal{Z}}$ is significantly smoother for learning and sampling.
\end{enumerate}

\begin{figure*}[t]
    \centering
    \includegraphics[width=\linewidth]{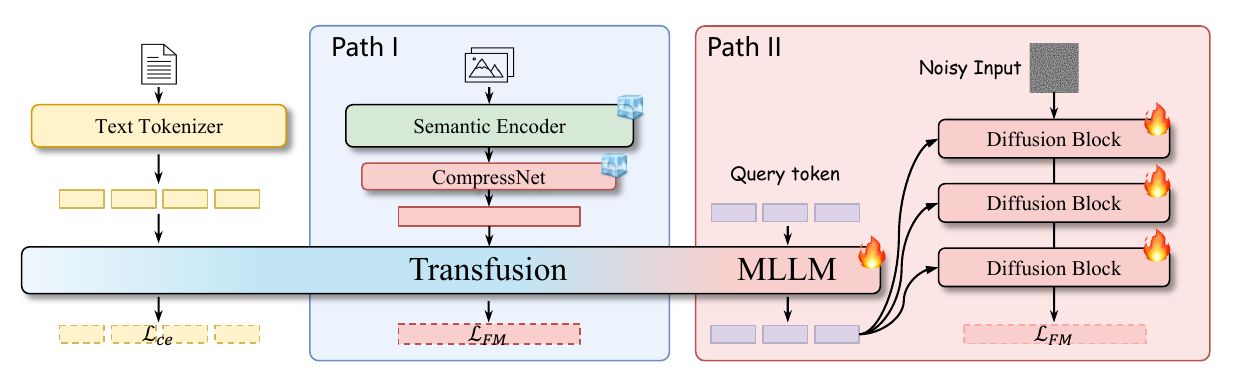}
    \caption{\textbf{Overview of the proposed framework.} For a controlled comparison, both pathways are built upon the same compressed representations and jointly optimized with cross-entropy loss ($\mathcal{L}_{ce}$) and flow matching loss ($\mathcal{L}_{fm}$).}
    \label{fig:unicom}
    \vspace{-1em}
\end{figure*}

Our framework, illustrated in Fig~\ref{fig:unicom}, operationalizes this formulation through three core components: Semantic Compressor \S\ref{compressor}, Generative Prior Module \S\ref{sec:two_ways}, and Diffusion Decoder. The compressor and decoder are first jointly pre-trained using a reconstruction objective to establish a well-behaved $\tilde{\mathcal{Z}}$ space. Subsequently, the compressor is frozen, and the Generative Prior Module is trained to sample within this fixed, tractable space.

\subsection{Semantic Compression and Reconstruction}
\label{compressor}
The core of our framework is the construction of the compressed semantic latent space $\tilde{\mathcal{Z}}$. This module must reconcile two competing objectives: \textbf{minimizing information loss} to preserve the visual and semantic fidelity required for high-quality reconstruction, and \textbf{maximizing tractability} by simplifying the underlying data distribution for robust generative modeling. Formally, we seek a parametric function $\mathcal{C}_{\phi} : \mathcal{Z} \rightarrow \tilde{\mathcal{Z}}$ that performs a nonlinear dimensionality reduction, where $\tilde{\mathcal{Z}} \subset \mathbb{R}^{N \times d}$ and $d \ll D$.

\paragraph{Why Compression is Necessary.}
The native VLM embedding space $\mathcal{Z}$, while semantically rich, exhibits high dimensionality and a complex, often multi-modal and discontinuous, topology. Directly modeling the distribution $P(\mathbf{Z} \mid \mathbf{c})$ with a generative model (\textit{e.g.}, a diffusion process) is empirically unstable and leads to high FID (see Tab~\ref{tab:recon_performance}), as the model struggles to navigate the intricate geometry of $\mathcal{Z}$. This observation aligns with the manifold hypothesis; we posit that the information essential for generation lies on a lower-dimensional submanifold within $\mathcal{Z}$. The compressor $\mathcal{C}_{\phi}$ thus acts as a learned projection that discovers and parameterizes this submanifold.

\paragraph{Design of the Attention-Based Compressor.}
Prior work, such as VUGEN~\cite{chen2025vugen}, employs a simple Multi-Layer Perceptron (MLP) for compression. While effective, an MLP processes each spatial feature token independently, potentially disregarding the crucial long-range contextual relationships between image patches that are inherently captured by the vision encoder's self-attention layers. To preserve this structured semantic information, we implement $\mathcal{C}_{\phi}$ as a shallow, lightweight Transformer module. This design ensures that the mapping is context-aware and permutation-equivariant with respect to the input tokens, preserving the structural semantics of the scene more effectively than an isotropic MLP (see Tab~\ref{tab:vqa_bench_ablation} and Fig~\ref{fig:tsne}).

\paragraph{Joint Optimization with the Diffusion Decoder.}
The compressor is not trained in isolation. Its parameters $\phi$ are jointly optimized with the diffusion decoder parameters $\psi$ using a reconstruction objective. Given an image $\mathbf{x}$, its visual features $\mathbf{Z} = f_{\text{und}}(\mathbf{x})$, and compressed latents $\tilde{\mathbf{z}} = \mathcal{C}_{\phi}(\mathbf{Z})$, the decoder reconstructs the image as $\hat{\mathbf{x}} = \mathcal{D}_{\psi}(\tilde{\mathbf{z}})$. Training minimizes the composite loss:
\begin{equation}
\mathcal{L}_{\text{recon}} =
\mathcal{L}_{\text{flow}}(\mathbf{x}, \hat{\mathbf{x}})
+ \lambda \cdot \mathcal{L}_{\text{perc}}(\mathbf{x}, \hat{\mathbf{x}}),
\end{equation}
where $\mathcal{L}_{\text{flow}}$ denotes the flow-matching loss used by the diffusion decoder and $\mathcal{L}_{\text{perc}}$ is a perceptual loss (e.g., LPIPS). This joint training procedure forces $\mathcal{C}_{\phi}$ to discard information redundant for pixel-space reconstruction while retaining semantically meaningful and generatively useful signals, thereby shaping $\tilde{\mathcal{Z}}$ into an \emph{information bottleneck} optimized for generation.

\subsection{Two Ways of Predicting Representations}
\label{sec:two_ways}
With the compressed semantic space $\tilde{\mathcal{Z}}$ established in~\Cref{compressor}, we explore two distinct pathways to model the conditional distribution $P(\tilde{\mathbf{z}} \mid \mathbf{c})$. Both pathways share the same compressor $\mathcal{C}_{\phi}$ and diffusion decoder $\mathcal{D}_{\psi}$, ensuring a controlled comparison focused on the prediction mechanism itself.

\paragraph{Pathway I: Unified Prediction with Transfusion.}
\label{transfusion}

This pathway integrates text and image generation within a single, fully trainable transformer model. The model operates on a unified sequence that interleaves discrete text tokens and continuous latent representations of images. Formally, given a text prompt $\mathbf{c}$, it is tokenized into a sequence of tokens $\{\mathbf{w}_1, \dots, \mathbf{w}_L\}$.    To generate an image, we prepend a special \texttt{[BOI]} (Beginning of Image) token and append an \texttt{[EOI]} (End of Image) token to a placeholder sequence of $N$ latent vectors initialized as pure noise $\tilde{\mathbf{z}}_T \sim \mathcal{N}(0, I)$. For visual inputs, the semantic encoder extracts continuous features $\mathbf{Z} = f_{\text{und}}(\mathbf{x})$. We directly input the uncompressed $\mathbf{Z}$ for understanding tasks to preserve fine-grained details, whereas for image editing, we feed the compressed latents $\tilde{\mathbf{z}} = \mathcal{C}_{\phi}(\mathbf{Z})$ to reduce context length and facilitate efficient multi-turn editing.

The transformer processes this entire sequence using a modality-aware attention mask.    For text tokens, we apply standard causal masking to preserve autoregressive properties.    For the latent vectors within the same image (between \texttt{[BOI]} and \texttt{[EOI]}), we employ \textit{bidirectional attention}, allowing all image patches to attend to each other.

\paragraph{Pathway II: Query-Guided Prediction via an MLLM.}
\label{metaquery}

This pathway decouples the understanding and generation roles. We employ a powerful pre-trained MLLM (\textit{e.g.}, Qwen-VL) with its multimodal reasoning and knowledge representation capabilities.    
To bridge it with the generative diffusion decoder, we introduce a set of learnable parameters called \textbf{MetaQueries}, denoted as $\mathcal{Q} \in \mathbb{R}^{M \times d}$, where $M$ is the number of queries and $d$ matches the MLLM's hidden dimension.

Given a text condition $\mathbf{c}$, we construct the input sequence as the concatenation of the text tokens and the MetaQueries: $[\mathbf{c};    \mathcal{Q}]$.    This sequence is fed into the frozen MLLM.    Through its forward pass, the queries $\mathcal{Q}$ interact with the text context via the MLLM's self-attention layers, effectively "querying" the model's internal knowledge and semantic understanding to produce a conditioned representation.    We extract the output states corresponding to the query positions as the conditional signal $\mathbf{h}_{\mathcal{Q}} \in \mathbb{R}^{M \times d}$.

This signal is then projected by a lightweight, trainable connector network (a small transformer encoder) to align it with the input space of the flow-matching decoder, producing the initial condition for the latent representation.

\paragraph{Flow Matching Training Objective.}
\label{sec:training}
During training, given a text condition $\mathbf{c}$ and its corresponding ground-truth compressed semantic representation $\tilde{\mathbf{z}}_1$, we construct training pairs following the standard Flow Matching procedure. Specifically, we sample a time step $t \sim \mathcal{U}[0,1]$ and noise $\epsilon \sim \mathcal{N}(0, I)$, and compute the interpolated latent
\begin{equation}
\tilde{\mathbf{z}}_t = t \tilde{\mathbf{z}}_1 + (1 - t)\epsilon,
\end{equation}
along with the target velocity
\begin{equation}
\mathbf{v}_t = \tilde{\mathbf{z}}_1 - \epsilon.
\end{equation}

The condition $\mathbf{c}$ and the noised latent $\tilde{\mathbf{z}}_t$ are assembled into the mixed input sequence and passed through the Transformer. The model predicts the velocity field from the output hidden states corresponding to the positions of $\tilde{\mathbf{z}}_t$. Optimization is performed using a mean squared error loss:
\begin{equation}
\mathcal{L}_{\text{FM}} =
\mathbb{E}_{t, \mathbf{c}, \tilde{\mathbf{z}}_1, \epsilon}
\left[
\left\|
\mathbf{v}_t - \mathbf{v}_{\theta}(\tilde{\mathbf{z}}_t, t; \mathbf{c})
\right\|_2^2
\right].
\end{equation}

During this stage, both the semantic compressor $\mathcal{C}_{\phi}$ and the diffusion decoder $\mathcal{D}_{\psi}$ are frozen. Consequently, the transformer learns to perform conditional sampling on a fixed, well-structured semantic manifold $\tilde{\mathcal{Z}}$.

\section{Experiments}
\begin{wraptable}{r}{0.5\textwidth}
  \centering
  \vspace{-1.5em}
  \caption{\textbf{Evaluation on image reconstruction ability on the ImageNet validation set}. $d$ denotes the feature dimension of compressed representations $\tilde{\mathbf{z}}$. Models marked with $^{\dagger}$ are re-evaluated using official checkpoints.}
  \label{tab:image-recon}
  \scriptsize
  \setlength{\tabcolsep}{3pt}
  \renewcommand{\arraystretch}{0.95}
  \begin{tabular}{lcccc}
    \toprule
    \textbf{Tokenizer} & \textbf{Res.} & \textbf{rFID$\downarrow$} & \textbf{PSNR$\uparrow$} & \textbf{SSIM$\uparrow$} \\
    \midrule
    \multicolumn{5}{l}{\textit{Specialized tokenizers}} \\
    SD-VAE~\cite{esser2024scalingrectifiedflowtransformers} & 256 & 1.06 & 28.62 & 0.86 \\
    GigaTok~\cite{xiong2025gigatok} & 256 & 0.51 & 21.32 & 0.69 \\
    VA-VAE~\cite{yao2025vavae} & 256 & 0.26 & 28.59 & 0.80 \\
    DC-AE~\cite{chen2024dcae} & 512 & 0.22 & 26.15 & 0.71 \\
    MAE-Tok~\cite{chen2025maetok} & 512 & 0.62 & - & - \\
    TexTok~\cite{zha2025textok} & 512 & 0.73 & 24.45 & 0.66 \\
    FLUX.1[dev]-VAE$^{\dagger}$~\cite{flux} & 512 & 0.06 & 33.65 & 0.93 \\
    \midrule
    \multicolumn{5}{l}{\textit{Unified tokenizers}} \\
    UniTok~\cite{ma2025unitok} & 256 & 0.38 & - & - \\
    TokenFlow~\cite{qu2025tokenflow} & 384 & 0.63 & 22.77 & 0.73 \\
    X-Omni$^{\dagger}$~\cite{chen2024dcae} & 512 & 8.30 & 15.66 & 0.38 \\
    MingTok$^{\dagger}$~\cite{ai2025ming} & 512 & 0.53 & 23.49 & 0.61 \\
    \rowcolor{mygreen}
    \textbf{Ours (d64)} & 1024 & 0.42 & 22.28 & 0.61 \\
    \rowcolor{mygreen}
    \textbf{Ours (d1152)} & 1024 & 0.38 & 22.60 & 0.61 \\
    \bottomrule
  \end{tabular}
  \vspace{-6em}
\end{wraptable}
In this section, we present {\method}, a unified multimodal model via compressed continuous semantic representations. We adopt the Transfusion architecture (Pathway~I in~\Cref{sec:two_ways}) as our default design, which is validated to outperform the Query-Guided alternative in~\Cref{sec:discussion}. 
We describe the implementation details in~\Cref{sec:implementation} and report performance on reconstruction, text-to-image generation, and image editing benchmarks in~\Cref{sec:main_results}. In~\Cref{sec:ablation}, we ablate the key design choices of our compressed representation, including the optimal feature shape and projector (compressor and decompressor) architecture.

\subsection{Implementation Details}
\label{sec:implementation}
Our unified model is built upon Qwen-2.5-7B-Instruct~\cite{qwen2.5} for multimodal understanding and FLUX.1-dev~\cite{flux} for high-fidelity image generation, utilizing SigLIP2~\cite{tschannen2025siglip2multilingualvisionlanguage} as the semantic encoder. 
To effectively bridge the modality gap, we employ a multi-head self-attention compressor for feature compression and a two-layer MLP for visual projection. 
The training pipeline consists of four progressive stages: alignment, pre-training, continued training, and supervised fine-tuning, all adopting a native-resolution strategy. 
Please refer to \Cref{app:implementation_app} for comprehensive details regarding the decoder architecture, hyperparameter settings, data mixtures, and the ablation setup.

\subsection{Main Results}
\label{sec:main_results}
\begin{figure*}[ht]
    \centering
    \includegraphics[width=\linewidth]{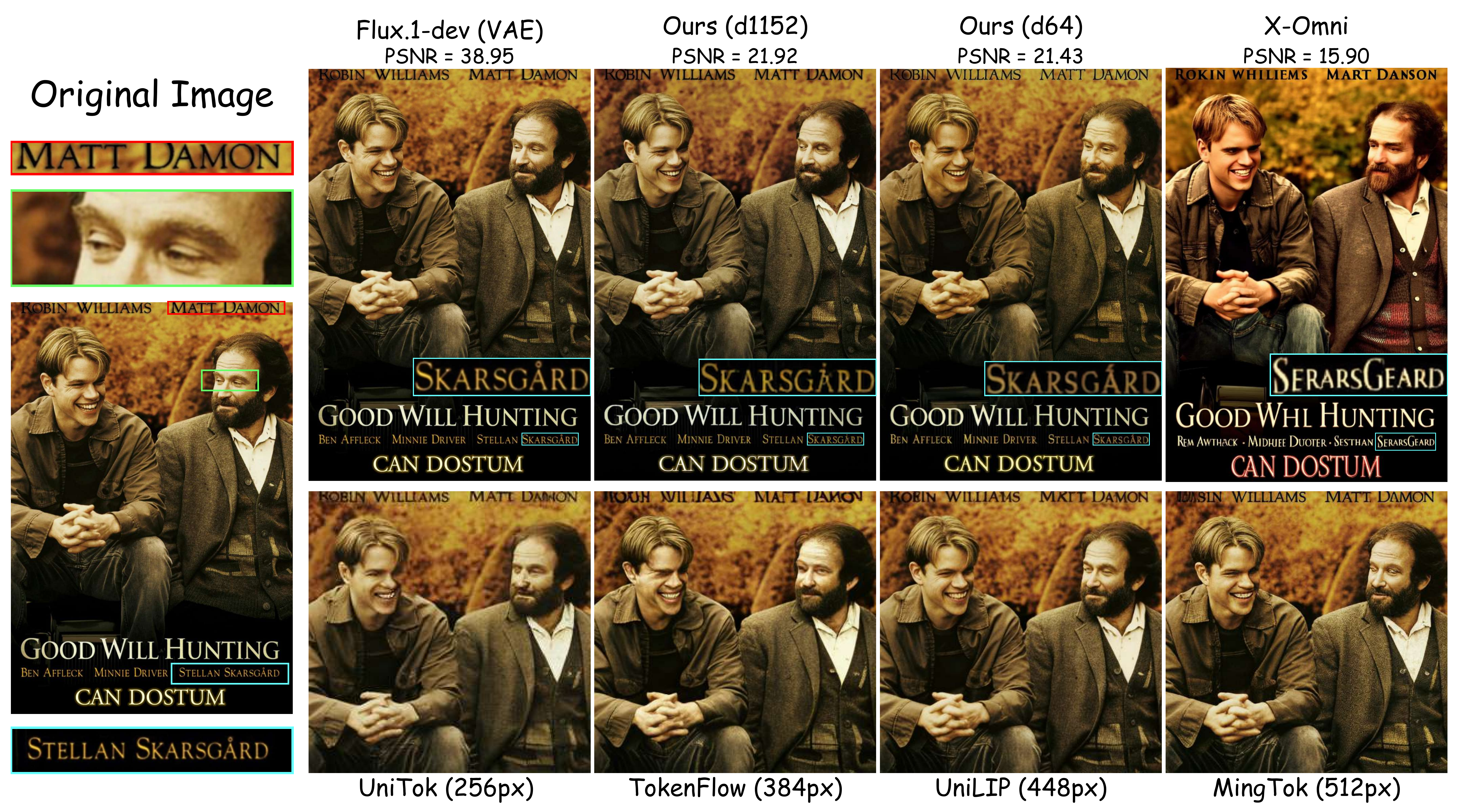}
    \vspace{-1.5em}
    \caption{\textbf{Visual comparison of image reconstruction results}. 
Notably, our method (d64) maintains high fidelity in high-frequency details (e.g., text characters) and preserves facial identity better than these semantic-based baselines, achieving quality comparable to the specialized Flux.1-dev VAE.
}
\label{fig:app_comparision_recon}
\vspace{-1.5em}
\end{figure*}

\paragraph{Diffusion Decoder.}
We report the quantitative evaluation on the ImageNet validation set~\cite{deng2009imagenet} in Tab~\ref{tab:image-recon}. Notably, comparing the uncompressed (d1152) and compressed (d64) variants, we observe that reducing the channel dimension by $18\times$ incurs negligible loss in reconstruction fidelity.
The qualitative superiority of our approach is further evidenced in Fig~\ref{fig:app_comparision_recon}. As shown in the zoomed-in regions, our method achieves reconstruction fidelity comparable to the specialized Flux VAE, effectively recovering high-frequency details such as small text characters and maintaining consistent facial identity. This stands in sharp contrast to prior methods relying on semantic features, which often exhibit severe degradation in these fine-grained structures.  Consistently with recent findings~\cite{tong2026scaling}, we observe that incorporating high-quality text-rich data during training is instrumental in enhancing text rendering performance. 

\begin{figure*}[htbp]
    \centering
    \includegraphics[width=\linewidth]{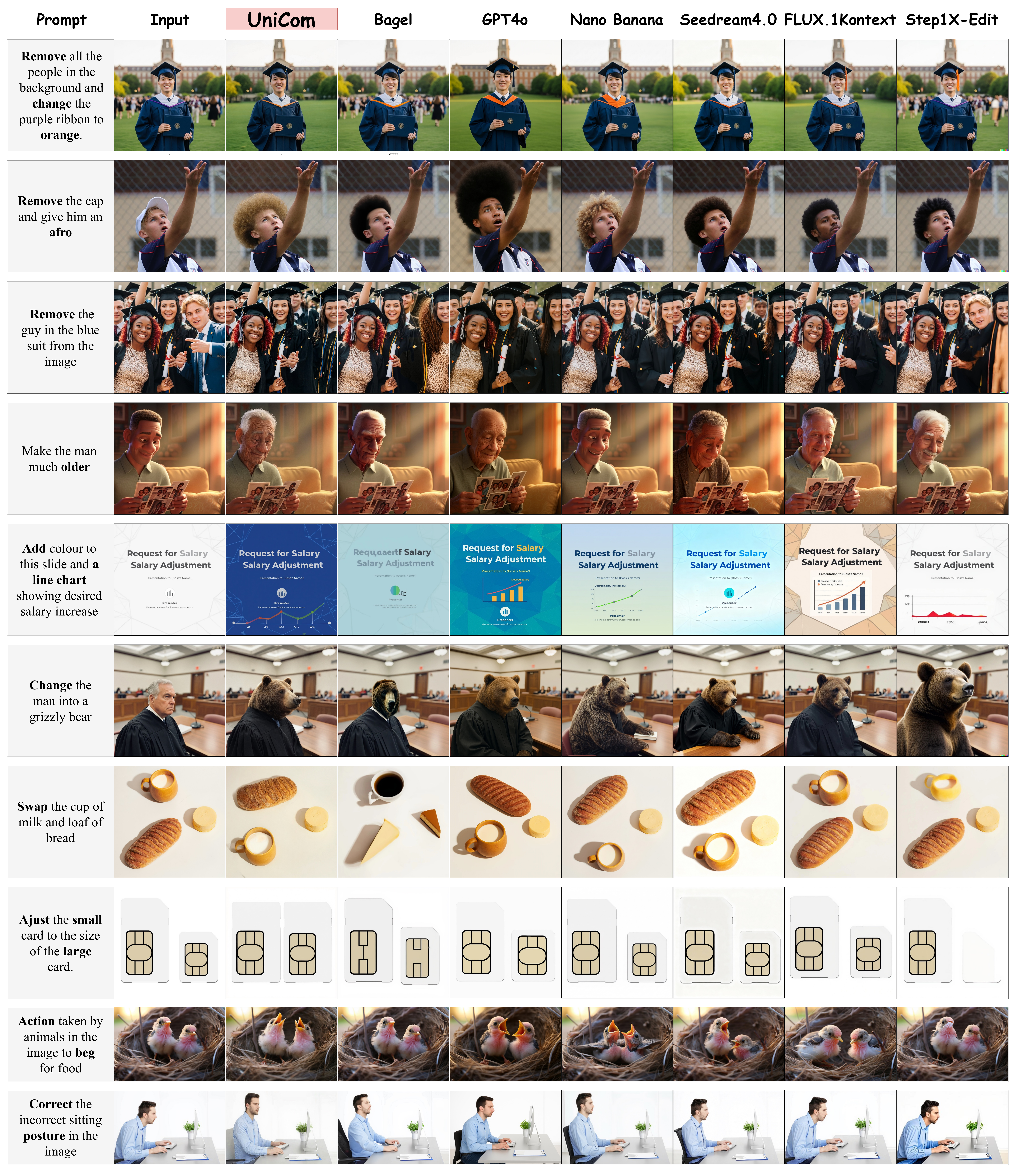}
    \caption{Comparison of the results for image editing, highlighting UniCom performance in tasks such as image manipulation, object swapping, and color adjustment. See more visualization in the Appendix.}
    \label{fig:placeholder}
    \vspace{-2em}
\end{figure*}

\begin{figure*}[t]
    \centering
    \includegraphics[width=\linewidth]{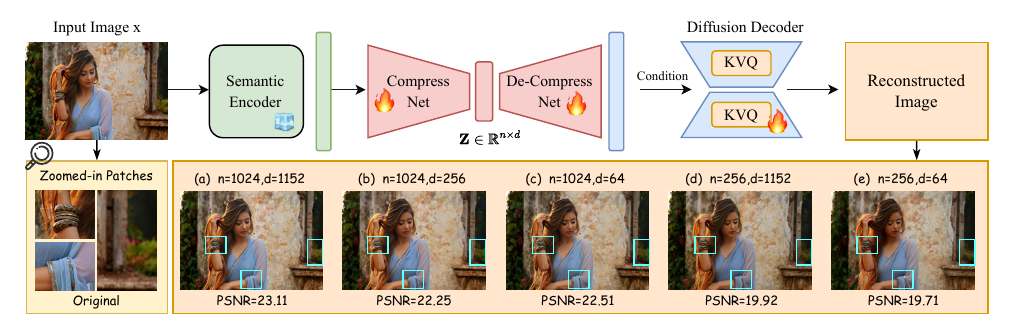}
    \vspace{-1em}
    \caption{\textbf{Overview of the proposed diffusion decoder and reconstruction analysis.} 
  Given an input image, the vision encoder extracts semantic features which are then processed by compressor and decompressor to condition the diffusion model for image reconstruction. 
During training, the compression modules and the diffusion backbone are optimized, while the encoder remains frozen. 
We investigate the bottleneck capacity by varying the token number $n$ and feature dimension $d$. 
As illustrated in the bottom row, compressing the dimension $d$ yields superior reconstruction fidelity, whereas reducing the token number $n$ leads to noticeable blurring in fine details.}
    \label{fig:diffusion_decoder}
\end{figure*}

\paragraph{Text to Image Generation.}
\begin{table*}[!t]
\centering
\caption{\textbf{Image generation results on GenEval, DPG-Bench, and WISE}. $^{\dagger}$ refers to methods using LLM rewriters on GenEval. Abbreviations for WISE: Cult. (Cultural), Bio. (Biology), Phy. (Physics), Chem. (Chemistry).}
\label{tab:merged_results}
\setlength\tabcolsep{1.8 pt} 
\resizebox{\linewidth}{!}{
\begin{tabular}{@{}l|ccccccc|c|ccccccc@{}}
\toprule
\multicolumn{1}{l|}{} & \multicolumn{7}{c|}{\textbf{GenEval}} & \multicolumn{1}{c|}{\textbf{DPG}} & \multicolumn{7}{c}{\textbf{WISE}} \\ \cmidrule(l){2-16} 
\multicolumn{1}{l|}{\multirow{-2}{*}{\textbf{Models}}} & \textbf{Single} & \textbf{Two} & \textbf{Count} & \textbf{Colors} & \textbf{Pos} & \textbf{Col-Attr} & \multicolumn{1}{c|}{\textbf{Overall}} & \textbf{Overall} & \textbf{Cult.} & \textbf{Time} & \textbf{Space} & \textbf{Bio.} & \textbf{Phy.} & \textbf{Chem.} & \textbf{Overall} \\ \midrule

\rowcolor{highlightgray} 
\multicolumn{16}{c}{\textit{Generation-only Models}} \\
\multicolumn{1}{l|}{SD3-Medium \citep{esser2024scaling}} & 0.99 & 0.94 & 0.72 & 0.89 & 0.33 & 0.60 & 0.74 & - & - & - & - & - & - & - & - \\
\multicolumn{1}{l|}{FLUX.1 [Dev] \citep{flux}} & 0.98 & 0.93 & 0.75 & 0.93 & 0.68 & 0.65 & 0.82 & 84.00 & 0.48 & \textbf{0.58} & 0.62 & 0.42 & 0.51 & 0.35 & 0.50 \\ \midrule
\rowcolor{highlightgray} 
\multicolumn{16}{c}{\textit{Unified Multimodal Models}} \\
\multicolumn{1}{l|}{MetaQuery-XL$^{\dagger}$\citep{pan2025transfer}} & - & - & - & - & - & - & 0.80 & - & \textbf{0.56} & {0.55} & 0.62 & \underline{0.49} & \underline{0.63} & \underline{0.41} & \underline{0.55} \\
\multicolumn{1}{l|}{Tar \citep{han2025vision}} & 0.99 & 0.92 & \underline{0.83} & 0.85 & 0.80 & 0.65 & 0.84 & 84.19 & - & - & - & - & - & - & - \\
\multicolumn{1}{l|}{BLIP3-o \citep{chen2025blip3}} & - & - & - & - & - & - & 0.84 & - & - & - & - & - & - & - & - \\
\multicolumn{1}{l|}{UniWorld-V1$^{\dagger}$ \citep{lin2025uniworld}} & 0.98 & 0.93 & 0.81 & 0.89 & 0.74 & 0.71 & 0.84 & - & 0.53 & {0.55} & \textbf{0.73} & {0.45} & 0.59 &\underline{0.41} & \underline{0.55} \\
\multicolumn{1}{l|}{OmniGen2$^{\dagger}$ \citep{wu2025omnigen2}} & 0.99 & \underline{0.96} & 0.74 & \textbf{0.98} & 0.71 & 0.75 & 0.86 & 83.57 & - & - & - & - & - & - & - \\
\multicolumn{1}{l|}{D-DiT \citep{li2025dual}} & 0.97 & 0.80 & 0.54 & 0.76 & 0.32 & 0.50 & 0.65 & - & - & - & - & - & - & - & - \\
\multicolumn{1}{l|}{Show-o \citep{xie2024show}} & 0.98 & 0.80 & 0.66 & 0.84 & 0.31 & 0.50 & 0.68 & - & 0.28 & 0.40 & 0.48 & 0.30 & 0.46 & 0.30 & 0.35 \\
\multicolumn{1}{l|}{Harmon \citep{wu2025harmonizing}} & 0.99 & 0.86 & 0.66 & 0.85 & 0.74 & 0.48 & 0.76 & - & 0.38 & 0.48 & 0.52 & 0.37 & 0.44 & 0.29 & 0.41 \\
\multicolumn{1}{l|}{MUSE-VL$^{\dagger}$\citep{xie2025muse}} & - & - & - & - & - & - & 0.57 & - & - & - & - & - & - & - & - \\
\multicolumn{1}{l|}{Transfusion \citep{zhou2024transfusion}} & - & - & - & - & - & - & 0.63 & - & - & - & - & - & - & - & - \\
\multicolumn{1}{l|}{Emu3 \citep{wang2024emu3}} & - & - & - & - & - & - & 0.66 & 81.60 & 0.34 & 0.45 & 0.48 & 0.41 & 0.45 & 0.27 & 0.39 \\
\multicolumn{1}{l|}{Show-o2 \citep{xie2025show}} & \textbf{1.00} & 0.87 & 0.58 & 0.92 & 0.52 & 0.62 & 0.76 & \underline{86.14} & - & - & - & - & - & - & - \\
\multicolumn{1}{l|}{Janus-Pro \citep{chen2025janus}} & 0.99 & 0.89 & 0.59 & 0.90 & 0.79 & 0.66 & 0.80 & 84.19 & 0.30 & 0.37 & 0.49 & 0.36 & 0.42 & 0.26 & 0.35 \\
\multicolumn{1}{l|}{Mogao \citep{liao2025mogao}} & \textbf{1.00} & \textbf{0.97} & \underline{0.83} & 0.93 & \underline{0.84} & \textbf{0.80} & \textbf{0.89} & 84.33 & - & - & - & - & - & - & - \\
\multicolumn{1}{l|}{X-Omni~\citep{geng2025x}} & 0.98 & 0.95 & 0.75 & 0.91 & 0.71 & 0.68 & 0.83 & \textbf{87.65} & - & - & - & - & - & - & - \\
\multicolumn{1}{l|}{Ming-UniVision~\citep{huang2025ming}} & \textbf{1.00} & 0.93 & 0.59 & 0.93 & \textbf{0.92} & 0.70 & 0.85 & 82.12 & - & - & - & - & - & - & - \\
\multicolumn{1}{l|}{BAGEL$^{\dagger}$ \citep{deng2025emerging}} & 0.98 & 0.95 & \textbf{0.84} & \underline{0.95} & 0.78 & \underline{0.77} & \underline{0.88} & 85.07 & 0.44 & {0.55} & {0.68} & 0.44 & 0.60 & 0.39 & 0.52 \\
\rowcolor{mygreen}
\multicolumn{1}{l|}{\textbf{UniCom (Ours)}} & 0.98 & 0.94 & {0.81}  & 0.91 & 0.82 & \underline{0.77} &  0.87 & 85.92 & \underline{0.55} & \underline{0.56} & \textbf{0.73} & \textbf{0.58} & \textbf{0.66} & \textbf{0.47} & \textbf{0.58} \\ 
\bottomrule
\end{tabular}}
\end{table*}

As shown in Tab~\ref{tab:merged_results}, on mainstream evaluation benchmarks including GenEval~\cite{ghosh2023geneval} and DPG-Bench~\cite{hu2024ella}, our method achieves performance comparable to state-of-the-art models. Particularly, UniCom achieves outstanding performance on the more challenging Wise~\cite{niu2025wise} bench. We attribute this advantage primarily to our method's direct use of visual-semantic features like SigLIP as the learning target. Compared to traditional VAE latent spaces, this semantically rich representation is inherently more suitable for encoding and reconstructing textual semantic information.

\paragraph{Image Editing.} 
As shown in Tab~\ref{tab:img_edit}, our model also demonstrates strong competitiveness in image editing tasks, achieving leading scores on both ImgEdit-Bench~\cite{ye2025imgedit} and GEdit-Bench~\cite{liu2025step1x-edit}. 
Many previous works~\cite{tong2026scaling, chen2025vugen, gao2025one} that generate based on comprehension priors either rarely systematically evaluate image editing capabilities, or still require introducing the reference image's VAE latent to maintain identity consistency~\cite{chen2025blip3o}. In contrast, our model conditions solely on the text instruction and the semantic features of the reference image, without using any VAE latent. Nonetheless, we still achieve high scores on metrics that measure consistency before and after editing. 
This strongly proves that our compressed representation can effectively preserve the fine-grained and structural information of the reference image. 
Notably, on more challenging knowledge-intensive benchmarks such as KRIS-Bench~\cite{wu2025kris} and WorldEdit~\cite{worldedit}, which require complex world knowledge spanning biology, geography, and cultural understanding, our model achieves the best performance among all open-source models. 
This demonstrates remarkable capability in handling sophisticated editing scenarios that demand deep semantic comprehension.

\begin{table*}[!t]
    \centering
    \caption{\textbf{Comparison of image editing capabilities.} We evaluate on ImgEdit-Bench, GEdit-Bench, KRIS-Bench and WorldEdit. For ImgEdit-Bench, performance is evaluated across nine distinct operation categories (e.g., `Add', `Adjust', `Extract', `Replace', `Remove', `Background', `Style', `Hybrid', and `Action'. ). For GEdit-Bench, metrics include `G-Semantic Consistency' (G-SC) and `G-Perceptual Quality' (G-PQ). For KRIS-Bench, we report Factual (Fact.), Conceptual (Conc.), and Procedural (Proc.) knowledge scores. \textbf{Bold}: best results. \underline{Underline}: second-best.}
    \label{tab:img_edit}
    \scriptsize
    \setlength\tabcolsep{3 pt}
    \resizebox{\linewidth}{!}{ 
    \begin{tabular}{@{}lccccccccccccc|ccc|c|c@{}}
    \toprule
    \multicolumn{1}{l|}{} & \multicolumn{10}{c|}{\textbf{ImgEdit-Bench}} & \multicolumn{3}{c|}{\textbf{GEdit-Bench}} & \multicolumn{4}{c|}{\textbf{KRIS-Bench}} & \multicolumn{1}{c}{\textbf{WorldEdit}} \\ 
    \cmidrule(l){2-19}
    \multicolumn{1}{l|}{\multirow{-2}{*}{\textbf{Models}}} & \textbf{Add} & \textbf{Adj.} & \textbf{Ext.} & \textbf{Rep.} & \textbf{Rm.} & \textbf{Bg.} & \textbf{Sty.} & \textbf{Hyb.} & \multicolumn{1}{c|}{\textbf{Act.}} & \multicolumn{1}{c|}{\textbf{Overall}} & \textbf{G-SC} & \multicolumn{1}{c|}{\textbf{G-PQ}} & \textbf{G-Overall} & \textbf{Fact.} & \textbf{Conc.} & \multicolumn{1}{c|}{\textbf{Proc.}} & \textbf{Overall} & \textbf{Overall} \\ \midrule

    \rowcolor{highlightgray} 
    \multicolumn{19}{c}{\textit{Generation-only Models}} \\
    \multicolumn{1}{l|}{FLUX.1 Kontext {[}Pro{]} \citep{batifol2025flux}} & 4.25 & 4.15 & 2.35 & 4.56 & 3.57 & 4.26 & 4.57 & 3.68 & \multicolumn{1}{c|}{4.63} & \multicolumn{1}{c|}{4.00} & 7.02 & \multicolumn{1}{c|}{\underline{7.60}} & 6.56 & 57.22 & 55.06 & \multicolumn{1}{c|}{46.69} & 54.17 & \underline{3.21} \\
    \multicolumn{1}{l|}{Qwen-Image \citep{wu2025qwen}} & \underline{4.38} & \underline{4.16} & \textbf{3.43} & \underline{4.66} & 4.14 & \underline{4.38} & \textbf{4.81} & \underline{3.82} & \multicolumn{1}{c|}{\textbf{4.69}} & \multicolumn{1}{c|}{\underline{4.27}} & \underline{8.00} & \multicolumn{1}{c|}{\textbf{7.86}} & \textbf{7.56} & - & - & \multicolumn{1}{c|}{-} & - & - \\ \midrule
    
    \rowcolor{highlightgray} 
    \multicolumn{19}{c}{\textit{Specialized Editing Models}} \\
    \multicolumn{1}{l|}{Instruct-Pix2Pix \citep{brooks2023instructpix2pix}} & 2.45 & 1.83 & 1.44 & 2.01 & 1.50 & 1.44 & 3.55 & 1.20 & \multicolumn{1}{c|}{1.46} & \multicolumn{1}{c|}{1.88} & 3.58 & \multicolumn{1}{c|}{5.49} & 3.68 & 23.33 & 25.59 & \multicolumn{1}{c|}{17.28} & 22.82 & 2.44 \\
    \multicolumn{1}{l|}{MagicBrush \citep{zhang2023magicbrush}} & 2.84 & 1.58 & 1.51 & 1.97 & 1.58 & 1.75 & 2.38 & 1.62 & \multicolumn{1}{c|}{1.22} & \multicolumn{1}{c|}{1.83} & 4.68 & \multicolumn{1}{c|}{5.66} & 4.52 & 41.84 & 39.24 & \multicolumn{1}{c|}{26.54} & 37.15 & 2.14 \\
    \multicolumn{1}{l|}{AnyEdit \citep{yu2025anyedit}} & 3.18 & 2.95 & 1.88 & 2.47 & 2.23 & 2.24 & 2.85 & 1.56 & \multicolumn{1}{c|}{2.65} & \multicolumn{1}{c|}{2.45} & 3.18 & \multicolumn{1}{c|}{5.82} & 3.21 & 39.26 & 41.88 & \multicolumn{1}{c|}{31.74} & 38.55 & 2.09 \\
    \multicolumn{1}{l|}{Step1X-Edit \citep{liu2025step1x-edit}} & 3.88 & 3.14 & 1.76 & 3.40 & 2.41 & 3.16 & 4.63 & 2.64 & \multicolumn{1}{c|}{2.52} & \multicolumn{1}{c|}{3.06} & 7.09 & \multicolumn{1}{c|}{6.76} & 6.70 & 45.52 & 48.01 & \multicolumn{1}{c|}{31.82} & 43.29 & - \\ \midrule
    
    \rowcolor{highlightgray} 
    \multicolumn{19}{c}{\textit{Unified Multimodal Models}} \\

    \multicolumn{1}{l|}{OmniGen \citep{xiao2025omnigen}} & 3.47 & 3.04 & 1.71 & 2.94 & 2.43 & 3.21 & 4.19 & 2.24 & \multicolumn{1}{c|}{3.38} & \multicolumn{1}{c|}{2.96} & 5.96 & \multicolumn{1}{c|}{5.89} & 5.06 & 33.11 & 28.02 & \multicolumn{1}{c|}{23.89} & 28.85 & 2.52 \\
    \multicolumn{1}{l|}{Ming-Univision \citep{huang2025ming}} & - & - & - &-  & - & - &-  & - & \multicolumn{1}{c|}{-} & \multicolumn{1}{c|}{-} & 6.04 & \multicolumn{1}{c|}{6.86} & 5.54 & - & - & \multicolumn{1}{c|}{-} & - & - \\
    \multicolumn{1}{l|}{BAGEL \citep{deng2025emerging}} & 3.56 & 3.31 & 1.70 & 3.30 & 2.62 & 3.24 & 4.49 & 2.38 & \multicolumn{1}{c|}{4.17} & \multicolumn{1}{c|}{3.20} & 7.36 & \multicolumn{1}{c|}{6.83} & 6.52 & \underline{60.26} & \underline{55.86} & \multicolumn{1}{c|}{\underline{51.69}} & \underline{56.21} & 2.76 \\
    \multicolumn{1}{l|}{UniWorld-V1 \citep{lin2025uniworld}} & 3.82 & 3.64 & 2.27 & 3.47 & 3.24 & 2.99 & 4.21 & 2.96 & \multicolumn{1}{c|}{2.74} & \multicolumn{1}{c|}{3.26} & 4.93 & \multicolumn{1}{c|}{7.43} & 4.85 & - & - & \multicolumn{1}{c|}{-} & - & - \\
    \multicolumn{1}{l|}{OmniGen2 \citep{wu2025omnigen2}} & 3.57 & 3.06 & 1.77 & 3.74 & 3.20 & 3.57 & \textbf{4.81} & 2.52 & \multicolumn{1}{c|}{\underline{4.68}} & \multicolumn{1}{c|}{3.44} & 7.16 & \multicolumn{1}{c|}{6.77} & 6.41 & 57.36 & 44.20 & \multicolumn{1}{c|}{47.79} & 49.71 & 2.51 \\ 
     \multicolumn{1}{l|}{TUNA~\citep{liu2025tuna}} & \textbf{4.46} & \textbf{4.52} & 2.47 & \textbf{4.68} & \textbf{4.58} & \textbf{4.56} & 4.73 & \textbf{4.07} & \multicolumn{1}{c|}{\textbf{4.69}} & \multicolumn{1}{c|}{\textbf{4.31}} & 7.79 & \multicolumn{1}{c|}{7.48} & 7.29 & - & - & - & - & - \\
    \rowcolor{mygreen}
    \multicolumn{1}{l|}{\textbf{UniCom (Ours)}} & 4.36  & 4.04 & \underline{3.30}  & 4.63 & \underline{4.40} & 4.24 & \underline{4.79} & 3.54 & \multicolumn{1}{c|}{\textbf{4.69}} & \multicolumn{1}{c|}{4.22} & \textbf{8.06} & \multicolumn{1}{c|}{7.33} & \underline{7.32} & \textbf{74.63} & \textbf{69.48} & \multicolumn{1}{c|}{\textbf{65.30}} & \textbf{70.11} & \textbf{4.12} \\
    \bottomrule
    \end{tabular}}
    \end{table*}

\subsection{Ablation Studies} 
\label{sec:ablation}
In this section, we rigorously dissect the design choices for the continuous unified representation $\tilde{\mathbf{z}}$. We select SigLIP2-SO400M-Patch16-NaFlex~\cite{tschannen2025siglip2multilingualvisionlanguage} features as the foundation due to their dense features for improved semantic understanding. However, directly modeling such high-dimensional representations is challenging. Therefore, we investigate how to efficiently compress these features from two perspectives: (1) Optimal Feature Shape (sequence vs. dimension compression), and (2) Projector Architecture (mlp vs. attention).

\subsubsection{Optimal Feature Shape}
We first determine the optimal shape for the compressed representation space $\tilde{\mathcal{Z}}$ by varying the sequence length $n$ and feature dimension $d$. 
We control the sequence length $n$ via the maximum token count and the dimension $d$ via a learnable mlp bottleneck (compressor and decompressor). 

\textbf{Reducing dimensions maintains higher reconstruction fidelity than reducing sequence length.}
As shown in Tab~\ref{tab:recon_performance} and Fig~\ref{fig:diffusion_decoder}, original siglip features yield excellent reconstruction fidelity (rFID 0.40). However, reduce sequence length drastically undermines image quality. In contrast, dimension reduction via MLP, whether compressing to 256 or 64 dimensions, results in relatively minimal performance loss.

\begin{figure*}[t]
  \centering
  \begin{minipage}[t]{0.48\textwidth}
    \centering
    \includegraphics[width=0.49\linewidth]{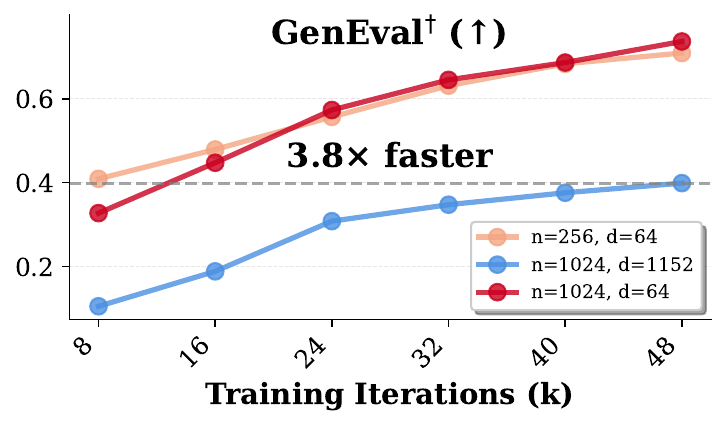}
    \hfill
    \includegraphics[width=0.49\linewidth]{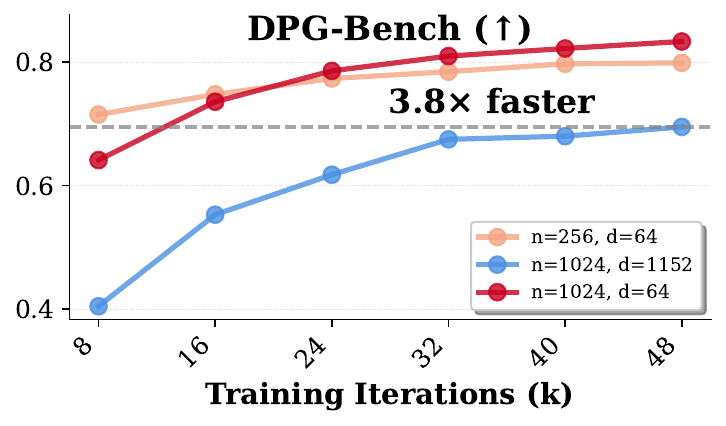}
    \captionof{figure}{\textbf{Feature dimension compression accelerates T2I training.} Compressing to $d=64$ achieves $3.8\times$ faster convergence than original SigLIP features.
}
    \label{fig:geneval_dpg_geneval_curves}
  \end{minipage}
  \hfill
  \begin{minipage}[t]{0.48\textwidth}
    \centering
    \includegraphics[width=0.49\linewidth]{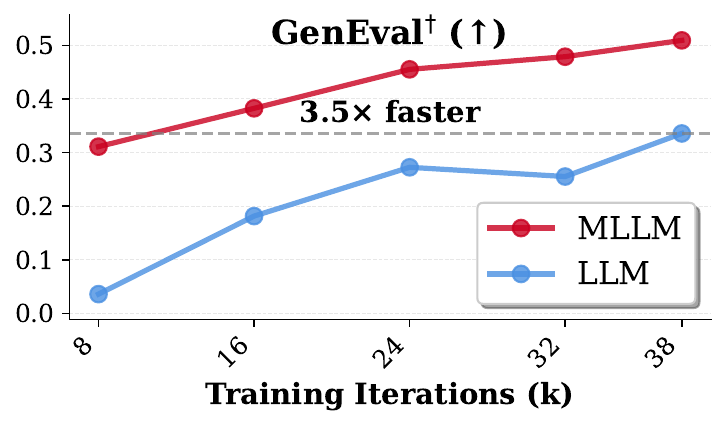}
    \hfill
    \includegraphics[width=0.49\linewidth]{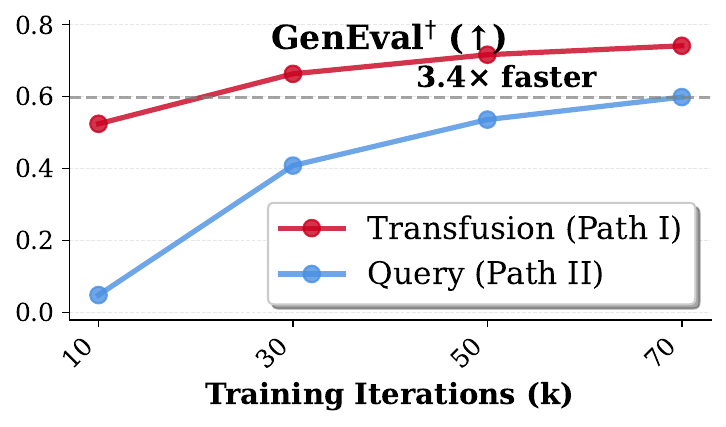}
    \captionof{figure}{
    \textbf{VLM initialization and Transfusion architecture improve generation.} We report GenEval results during unify pre-training.}
    \label{fig:aba_disc}
  \end{minipage}
  \vspace{-1em}
\end{figure*}

\textbf{Reducing dimensions yields superior image generation performance than reducing sequence length.}
\begin{wraptable}{r}{0.4\textwidth}
  \vspace{-1em}
  \centering
  \caption{Reconstruction performance of different compressed feature shapes on ImageNet50k~\citep{deng2009imagenet}.}
  \label{tab:recon_performance}
  \scriptsize
  \setlength{\tabcolsep}{3pt}
  \begin{tabular}{l c c c c c}
    \toprule
    \multirow{2}{*}{\textbf{Projector}} & \multirow{2}{*}{\textbf{Tokens ($n$)}} & \multirow{2}{*}{\textbf{Dim ($d$)}} & \multicolumn{3}{c}{\textbf{Reconstruction}} \\
    \cmidrule(l){4-6} 
     & & & rFID $\downarrow$ & PSNR $\uparrow$ & SSIM $\uparrow$ \\
    \midrule
    -- & 1024 & 1152 & 0.40 & 23.26 & 0.69 \\
    -- & 256 & 1152 & 0.72 & 20.29 & 0.56 \\
    MLP & 1024 & 256 & 0.62 & 21.73 & 0.66 \\
    MLP & 1024 & 64 & \textbf{0.55} & 22.17 & 0.66 \\
    MHA & 1024 & 64 & 0.56 & \textbf{22.61} & \textbf{0.69} \\
    \bottomrule
  \end{tabular}
  \vspace{-2em}
\end{wraptable}
Beyond pixel-level reconstruction, we assess how feature shapes impact the generative modeling dynamics of our unified model. As shown in Fig~\ref{fig:geneval_dpg_geneval_curves}, directly modeling uncompressed high-dimensional features ($d=1152$) results in slow convergence and suboptimal performance. 
While predicting such high-dimensional targets often requires specific optimization tricks~\cite{zheng2025diffusion}, these methods may not scale effectively across massive multi-task data. 
In contrast, compressing the dimension to $d=64$ accelerates convergence speed by approximately $5\times$ and achieves higher final quality. 
We further observe that maintaining the full token sequence ($n=1024$) exhibits greater potential in later training stages compared to the token-reduced variant ($n=256$). 
Moreover, the high reconstruction fidelity from the full sequence is essential for handling complex editing tasks. 
Consequently, we select the channel-compressed configuration ($n=1024, d=64$) as our optimal training target.

\subsubsection{Projector Architecture}
We compare a simple \textbf{MLP} module against a Multi-Head Attention (\textbf{MHA}) projector inspired by Unitok~\cite{ma2025unitok}. From a reconstruction perspective (Fig~\ref{tab:recon_performance}), both MLP and MHA projectors yield comparable fidelity. However, as our goal is a unified representation, the compressed representations must also retain rich semantic information. We validate this capability through two approaches: feature distribution visualization and downstream understanding benchmarks.

\begin{figure}
    \centering
    \includegraphics[width=\linewidth]{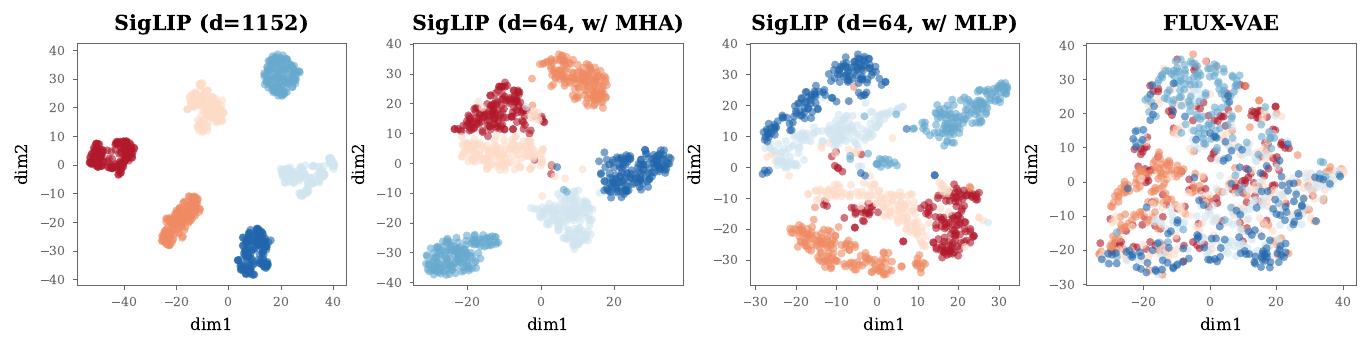}
    \caption{\textbf{T-SNE visualization of feature distributions.} 
We randomly select 6 distinct classes from ImageNet, sampling 150 images per class. 
MHA compression to 64-d preserves the structure, MLP projection yields scattered 64-d features, and FLUX-VAE has no semantic separability due to pixel-reconstruction focus.
}
    \label{fig:tsne}
    \vspace{-1em}
\end{figure}

\textbf{Distributional Analysis (t-SNE).} 
We visualize the feature distributions of different encoding strategies in Fig~\ref{fig:tsne}.
As shown in Fig~\ref{fig:tsne}, the MHA projector produces tightly clustered embeddings that maintain the structured distribution of the original SigLIP2 feature. In contrast, the MLP projection results in scattered clusters with blurred semantic boundaries. This suggests that the attention mechanism effectively leverages intra-token context to preserve high-level semantic alignment, whereas the MLP treats tokens in isolation, leading to semantic degradation.

\begin{wraptable}{r}{0.5\textwidth}
\centering
\caption{
Visual understanding performance on six benchmarks using compressed representations as visual inputs.
}
\label{tab:vqa_bench_ablation}
\scriptsize
\small
\setlength{\tabcolsep}{4pt} 
\resizebox{\linewidth}{!}{
\begin{tabular}{l | ccc c cc}
\toprule
\multirow{2}{*}{\textbf{Model Config.}} & \multicolumn{3}{c}{\textbf{General VQA}} & \textbf{Knowledge} & \multicolumn{2}{c}{\textbf{Text-Centric}} \\
\cmidrule(lr){2-4} \cmidrule(lr){5-5} \cmidrule(l){6-7}
 & GQA & RealWorldQA & SEED & MMMU & ChartQA & OCRBench \\
\midrule
\rowcolor{highlightgray} 
Baseline & \textbf{65.25} & 64.31 & \textbf{74.63} & 44.56 & 69.04 & 55.40 \\
\midrule
\multicolumn{7}{l}{\textit{Compression Strategies}} \\
\midrule
MLP Proj. & 62.80 & 60.39 & 69.92 & 43.00 & 56.80 & 31.70 \\
MHA Proj. & 64.01 & 63.14 & 71.75 & 44.11 & 62.12 & 36.00 \\
\midrule
\multicolumn{7}{l}{\textit{Concatenation Strategies}} \\
\midrule
Dim. Concat & 64.22 & 60.26 & 73.11 & \textbf{45.00} & 60.88 & 48.10 \\
Seq. Concat & 65.03 & \textbf{64.58} & 73.62 & 43.33 & \textbf{69.24} & \textbf{55.50} \\
\bottomrule
\end{tabular}
}
\vspace{-1em}
\end{wraptable}

\textbf{Downstream Visual Understanding Tasks.}
To quantify semantic capability, we employ the compressed representations $\tilde{\mathbf{z}}$ within a multimodal LLM framework~\cite{liu2023visual}, pre-training on LLaVA-Pretrain-557k~\cite{liu2023visual} and fine-tuning on Cambrian-737k~\cite{tong2024cambrian}. 
As detailed in Tab~\ref{tab:vqa_bench_ablation}, the MHA projector consistently outperforms the MLP, confirming that attention-based compression better preserves complex semantics. Although pure compression incurs minor information loss compared to the uncompressed baseline on text-rich tasks, we find that this can be effectively mitigated by fusing the MHA-compressed features with the original siglip features. 
Specifically, sequence concatenation of these two representations even introduces marginal gains on fine-grained benchmarks. Thus, we adopt the MHA projector as our final design.

\subsection{Discussion}
\label{sec:discussion}
\textbf{Visual Understanding Inform Image Generation.} 
To investigate whether the semantic knowledge acquired from understanding tasks contributes to generative performance, we compare two initialization strategies for unified training: 
(1) Language model from Qwen-2.5-7B~\cite{qwen2.5}, and 
(2) Pre-trained VLM, where the Qwen-2.5-7B (coupled with SigLIP2 features) was previously trained on image-to-text understanding tasks. 
Both models are subsequently trained on the unified generation task under identical settings. Fig~\ref{fig:aba_disc} (Left) compares their training dynamics on the generation task. The \textit{VLM-Base} demonstrates superior performance, characterized by significantly faster convergence and higher final metrics on GenEval and DPG-Bench. This indicates that dense visual-language alignment established during the understanding phase effectively bootstraps the generation process.

\textbf{Comparison with Query-Guided Architectures.} 
We also explored the Query-Guided pathway (Pathway~II), inspired by BLIP-3o~\cite{chen2025blip3}. The original design suffers from \textbf{feature misalignment}, as the frozen MLLM uses one vision encoder for understanding but is trained to predict features from a different encoder for generation. We addressed this by enforcing a unified feature space with SigLIP2, and further introduced an I2I reconstruction task inspired by RECA~\cite{xie2025reconstruction} to improve structural alignment.

Despite these improvements, Pathway~II remains inferior to Pathway~I. As shown in Fig~\ref{fig:aba_disc} (Right), the query-based model converges significantly slower during pre-training. Moreover, the query bottleneck discards spatial details, failing to preserve fine-grained layout. In contrast, Pathway~I benefits from full-sequence modeling with dense spatial correspondence, yielding superior fidelity and structural consistency in editing (see Fig~\ref{fig:query}).

\section{Conclusion}
In this paper, we introduce UniCom, a unified multimodal framework that effectively integrates visual understanding and generation by compressing high-dimensional visual embeddings into a compact latent space.  Based on the the continuous semantic compressor, we explore two distinct generative pathways: Transfusion and MetaQuery.
Experimental results show that UniCom not only preserves high-level semantics but also retains fine-grained visual details, achieving significant improvements in text-to-image generation and image editing tasks when compared to existing methods.  Furthermore, UniCom demonstrates that compressed semantic representations can be used for both unified generation and understanding without relying on variational autoencoders (VAE).
Future work could focus on further optimizing the compression module and expanding the application of this framework to other domains, such as video generation and multimodal reasoning.

\bibliography{main}
\bibliographystyle{iclr2025_conference}
\newpage
\appendix
\newpage
\appendix
\onecolumn

\section{Implementation details}
\label{app:implementation_app}
\subsection{Decoder Setup.} As illustrated in~\Cref{fig:diffusion_decoder}, we employ the SigLIP2-SO400M-Patch16-NaFlex~\cite{tschannen2025siglip2multilingualvisionlanguage} as our vision encoder. To enable high-fidelity image generation, we initialize the diffusion backbone with FLUX.1-dev~\cite{flux}. Crucially, to bridge the dimensionality gap between dense semantic features and the generative latent space, we implement a multi-head self-attention mechanism following UniTok~\cite{ma2025unitok}, which effectively preserves rich semantics during feature compression. 
We optimize both the compression module and the diffusion model on a high-quality internal dataset. We adopt a robust multi-resolution training strategy. Specifically, we define a grid of 33 discrete aspect ratio buckets, ranging from 1:4 (vertical) to 4:1 (horizontal), anchored at a base resolution of $1024 \times 1024$. Training is conducted with a global batch size of 256. We observe rapid convergence, achieving high-quality reconstruction performance within 10K steps. However, we extend training to a total of 50K steps to ensure the full recovery of fine-grained visual details.

\subsection{Unified Model Setup.} 
\begin{figure}[ht]
\centering
\begin{minipage}[c]{0.50\textwidth}
    \centering
    \includegraphics[width=\linewidth]{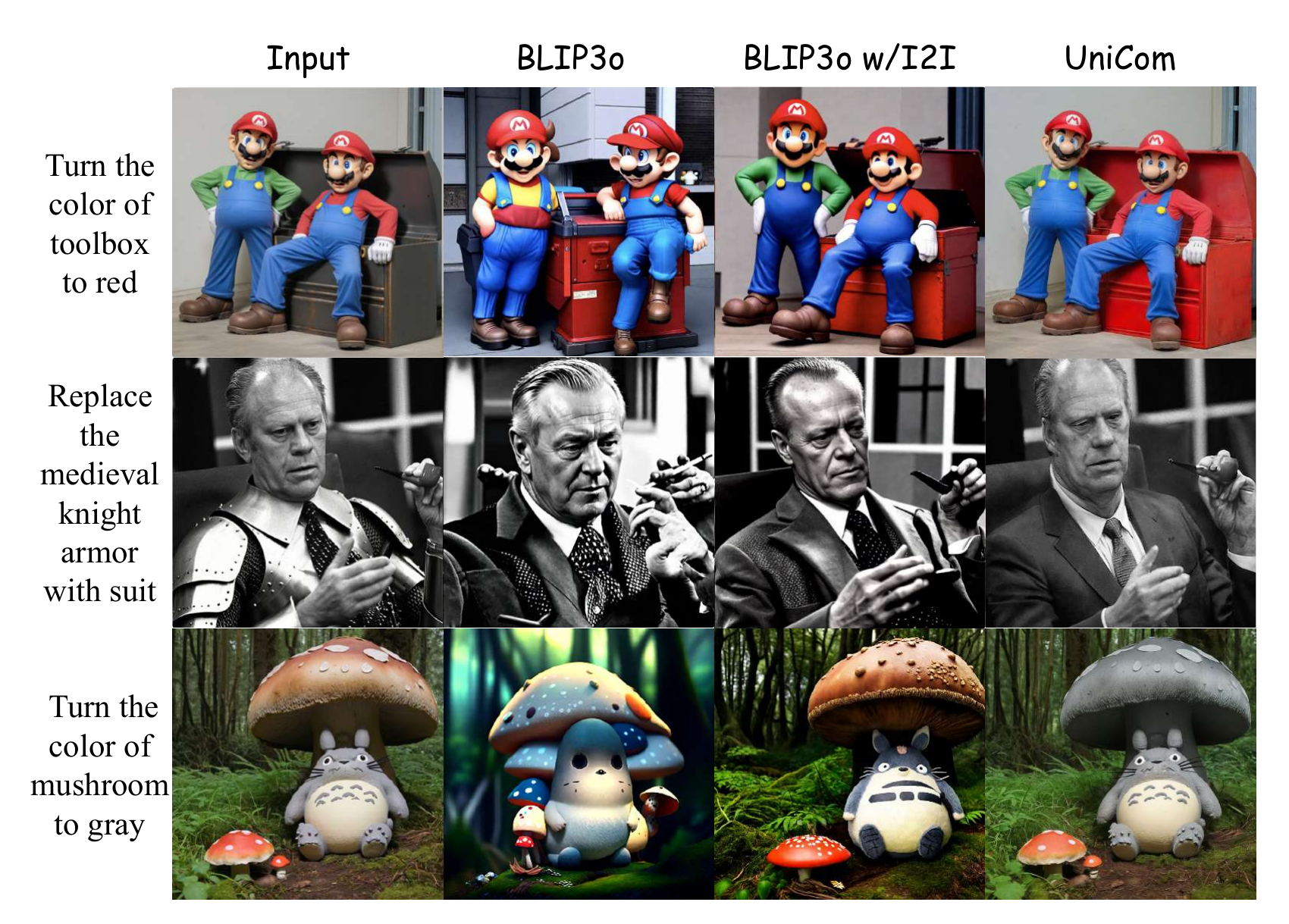}
    \captionof{figure}{Our unified prediction architecture (\textit{UniCom}) preserves finer spatial layout and structural consistency, while query-based methods struggle with spatial correspondence and editing fidelity.}
    \label{fig:query}
\end{minipage}
\hfill
\begin{minipage}[c]{0.47\textwidth}
    \centering
    \captionof{table}{\textbf{Training hyperparameters and datasets.}}
    \label{tab:data_mix}
    \renewcommand{\arraystretch}{1.15}
    \setlength{\tabcolsep}{3pt}
    \resizebox{\linewidth}{!}{
    \begin{tabular}{l | c c c c}
    \toprule
     & \textbf{Alignment} & \textbf{PT} & \textbf{CT} & \textbf{SFT} \\
    \midrule
    \multicolumn{5}{l}{\textbf{Hyperparameters}} \\
    \midrule
    Learning rate & $2.0 \times 10^{-5}$ & $1.0 \times 10^{-4}$ & $1.0 \times 10^{-4}$ & $1.0 \times 10^{-5}$ \\
    LR scheduler & Cosine & Constant & Constant & Constant \\
    Weight decay & 0.0 & 0.01 & 0.01 & 0.01 \\
    Gradient norm clip & 1.0 & 1.0 & 1.0 & 1.0 \\
    Optimizer & \multicolumn{4}{c}{AdamW ($\beta_1\!=\!0.9, \beta_2\!=\!0.95, \epsilon\!=\!10^{-6}$)} \\
    Loss weight ($\lambda_{\text{diff}}\!:\!\lambda_{\text{text}}$) & - & 5:1 & 5:1 & 5:1 \\
    Warm-up steps & 600 & 5000 & 5000 & 100 \\
    Training steps & 20K & 115K & 60K & 7K \\
    \midrule
    \multicolumn{5}{l}{\textbf{Data sampling ratio}} \\
    \midrule
    Text-only & 0.2 & 0.25 & 0.20 & 0.10 \\
    Image-Text (T2I) & 0.0 & 0.45 & 0.40 & 0.20 \\
    Image-Text (I2T) & 0.8 & 0.15 & 0.20 & 0.40 \\
    Image-Text-Image & 0.0 & 0.15 & 0.20 & 0.30 \\
    \bottomrule
    \end{tabular}
    }
\end{minipage}
\end{figure}

We initialize the language component with Qwen2.5-7B-Instruct~\cite{qwen2.5} and the vision encoder with SigLIP2-SO400M-Patch16-NaFlex~\cite{tschannen2025siglip2multilingualvisionlanguage}. Visual features are projected into the LLM's embedding space via a two-layer MLP projector. As outlined in~\Cref{tab:data_mix}, we employ a multi-stage training pipeline using a dynamic mixture of curated data.
Throughout all phases, we keep the SigLIP2 vision encoder frozen and adopt the same multi-resolution strategy utilized in the decoder setup, processing images with dynamic token counts up to a maximum of 1,024. The training process proceeds as follows:
(1) \textbf{Alignment}: We establish fundamental visual understanding capabilities by optimizing both the MLP connector and the language model parameters. 
(2) \textbf{Pre-training}: Subsequently, we inject generative capabilities using a massive corpus of text, image-text pairs, and image-text-image sequences. 
(3) \textbf{Continued Training}: To enhance performance on complex tasks such as inpainting and subject-driven generation, we strategically increase the sampling ratio of image-text-image pairs. 
(4) \textbf{Supervised Fine-tuning}: Finally, the model undergoes fine-tuning on high-quality curated data to ensure precise instruction following across all modalities.

\subsection{Ablation Setup.} 
For efficient ablation analysis in~\Cref{sec:ablation}, we switch to reduced data scales and simple task configurations. When investigating reconstruction fidelity, we train the compression and diffusion modules on a subset of high-quality internal images. Regarding generative modeling dynamics, we train the model on a filtered text-to-image dataset to assess convergence speed and generation quality. We evaluate the model's multimodal understanding capabilities on multiple benchmarks, including general VQA benchmarks such as GQA~\cite{hudson2019gqa}, RealWorldQA~\cite{xai2024-grok15v}, and SEED-Bench~\cite{li2023seed};  knowledge-intensive benchmarks such as MMMU~\citep{yue2024mmmu}; and text-centric benchmarks including ChartQA~\cite{masry2022chartqa}, OCRBench~\cite{liu2024ocrbench}.

\section{Related Work}
The visual tokenizer~\cite{pan2025generative,ma2025unitok,han2025vision} stands as a cornerstone in the realization of unified multimodal models. Its paramount function resides in transmuting visual imagery into visual tokens amenable to processing by generative models, thereby facilitating the concurrent support of both image understanding and generation tasks.
Traditional visual generation has long been predicated on pixel-based encoders rooted in Variational Autoencoders (VAEs~\cite{rombach2022high}), which affect image generation by projecting visual data into a low-dimensional latent space. Nevertheless, such pixel-centric encoders are often deficient in high-level semantic features and fail to adequately harness the priors inherent in visual understanding, a limitation that constrains their performance in generative endeavors.    
To remediate this shortcoming, several lines of research~\cite{chen2025janus,wang2025skywork,qu2025tokenflow,xie2025show,deng2025emerging} have advanced Hybrid Encoding methodologies, which integrate pixel encoding (VAE) with Transformer-based Vision Transformer (ViT) models in a bid to unify image generation and understanding.    Yet, this approach remains flawed: generation and understanding tasks still operate within disparate representational spaces, engendering a disjunction between the two and compromising overall efficiency.

More sophisticated paradigms pivot toward ViT-based visual tokenizers~\cite{geng2025x,ma2025unitok,han2025vision}. 
For instance, Vector Quantization (VQ)-based methods~\cite{esser2021taming} generate discrete visual tokens by quantizing continuous ViT features. While such approaches can effectively convey image information to generative models, the quantization process inevitably incurs information loss, and discrete token systems struggle to articulate the nuanced complexities of intricate visual features. 
By way of contrast, recent work has gravitated toward the employment of continuous ViT features for visual representation, exemplified by frameworks such as MingTok~\cite{huang2025ming} and VUGEN~\cite{chen2025vugen}. These methodologies eschew the discretization process; by adopting continuous visual representations, they succeed in unifying understanding and generation tasks within a shared representational space. They surmount the twin pitfalls of information loss and task disjunction that plagued traditional methods, thereby elevating both the quality and diversity of generated images.

\section{Impact of Dimensionality Compression on Training}
\begin{figure}[ht]
    \centering
    \includegraphics[width=1\linewidth]{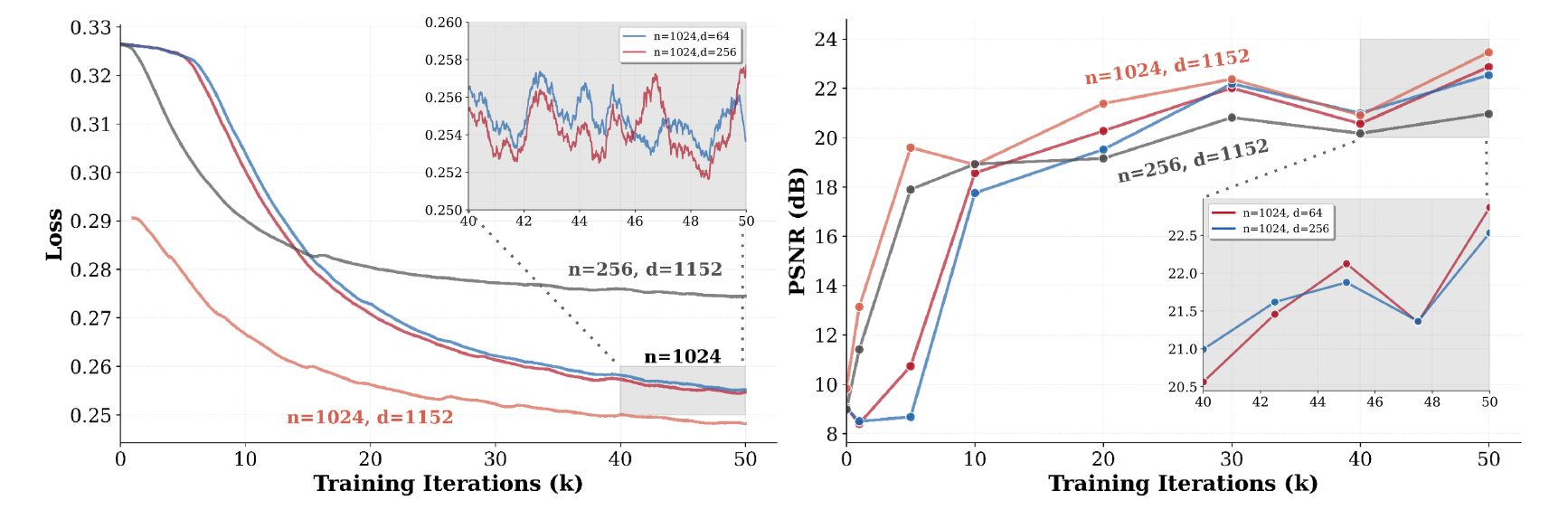}
    \caption{\textbf{Reducing dimensions maintains higher reconstruction fidelity than sequence reduction.} 
We compare the reconstruction training dynamics (Loss and PSNR) of different compressed representations ($\tilde{\mathbf{z}}$). }
    \label{fig:training_dynamics}
\end{figure}
\Cref{fig:training_dynamics} further illustrates that while dimension compression exhibits slower initial convergence compared to the sequence variant (step$<10K$), it rapidly matches the capability of the uncompressed baseline. Notably, we observe that as data and compute scale, the performance gap becomes negligible.
This suggests that the dense semantic features are highly compressible along the channel dimension without sacrificing pixel details.

\section{More Reconstruction Results}
\begin{figure}
    \centering
    \includegraphics[width=\linewidth]{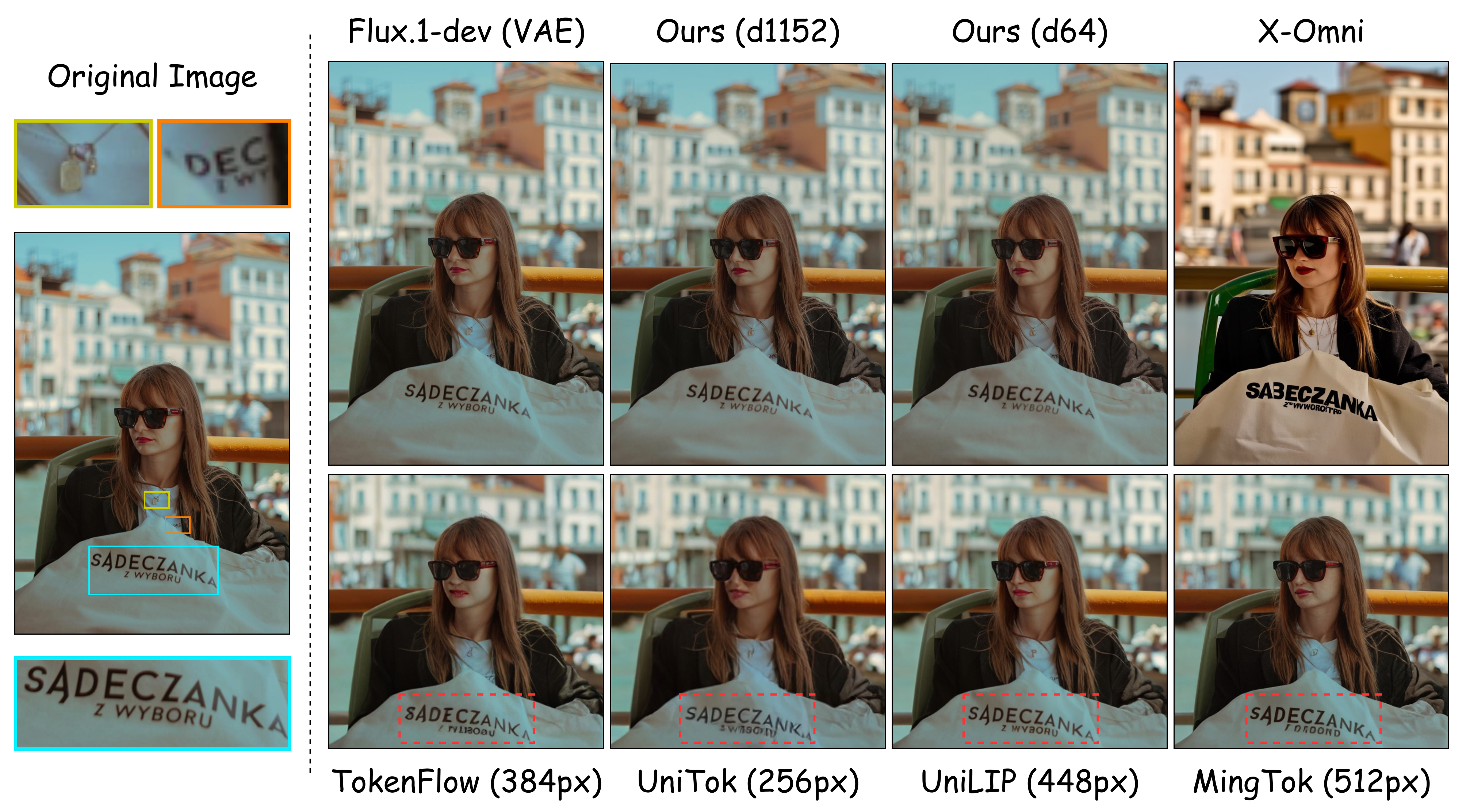}
    \caption{Additional visual comparison of image reconstruction results.}
    \label{fig:sota_compare_app}
\end{figure}

\begin{figure}
    \centering
    \includegraphics[width=\linewidth]{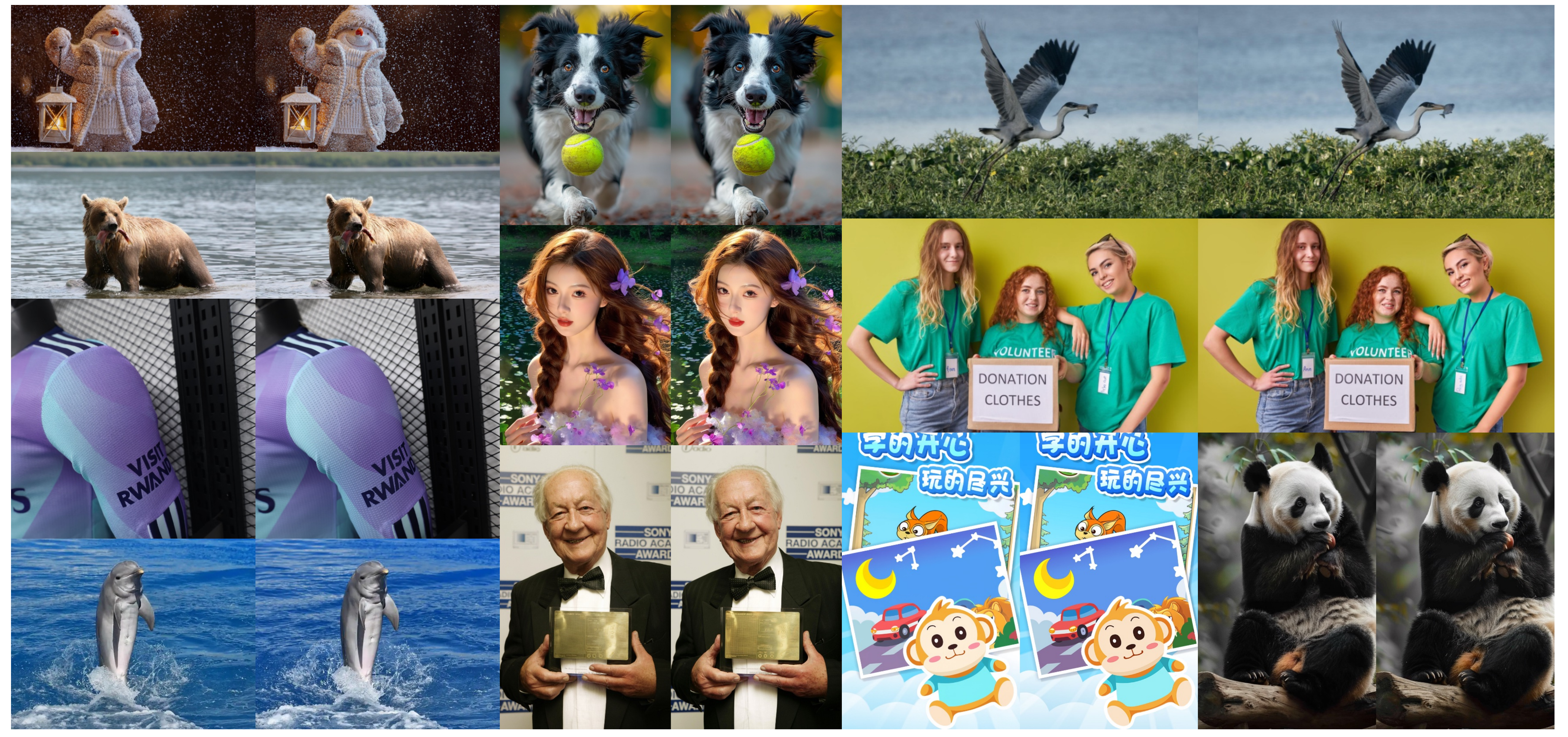}
    \caption{\textbf{Qualitative reconstruction samples using Decoder (d64).} 
  \textbf{Left:} Reconstructed image. \textbf{Right:} Original image. 
  Even with an $18\times$ channel compression ratio, our model accurately recovers complex high-frequency details.}
    \label{fig:more_cases_app}
\end{figure}

We provide comprehensive qualitative evaluations in ~\Cref{fig:sota_compare_app} and ~\Cref{fig:more_cases_app}. As illustrated in ~\Cref{fig:sota_compare_app}, prior unified tokenizers relying on semantic encoders often struggle with high-frequency details, resulting in blurred text and distorted small objects. In contrast, our approach preserves these fine-grained structures with high precision. Furthermore, ~\Cref{fig:more_cases_app} presents a direct side-by-side comparison between our channel-compressed reconstructions (d64) and the original images. The results demonstrate that our method retains intricate details, such as fabric textures, lighting gradients, and clothing lettering, rendering the reconstructed images perceptually indistinguishable from the ground truth. 

\section{More Results of Image Editing}
\label{app:editing_showcases}

\begin{figure*}[t]
    \centering
    \includegraphics[width=\linewidth]{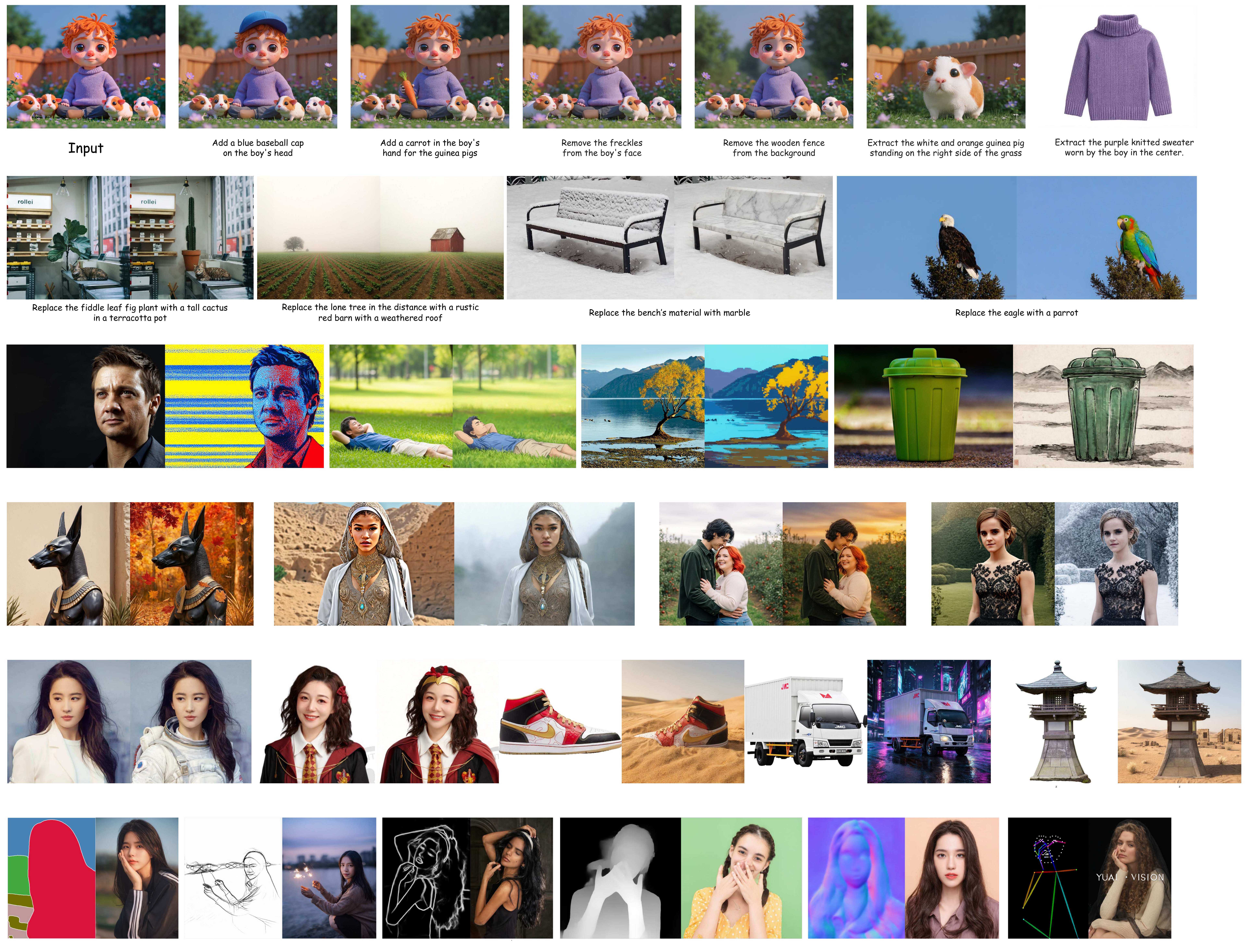}
    \caption{\textbf{More results on single image editing.} From top to bottom: (1) \textit{Add/Remove/Extract}: adding objects, removing elements, and extracting subjects; (2) \textit{Replace}: substituting objects or materials while preserving context; (3) \textit{Style Transfer}: transforming image appearance into different artistic styles; (4) \textit{Background}: modifying or replacing image backgrounds; (5) \textit{Subject-Driven Generation}: generating new images conditioned on reference subjects; (6) \textit{Controllable Generation}: generating images with spatial or structural control signals.}
    \label{fig:single_editing_app}
\end{figure*}

\begin{figure*}[t]
    \centering
    \includegraphics[width=\linewidth]{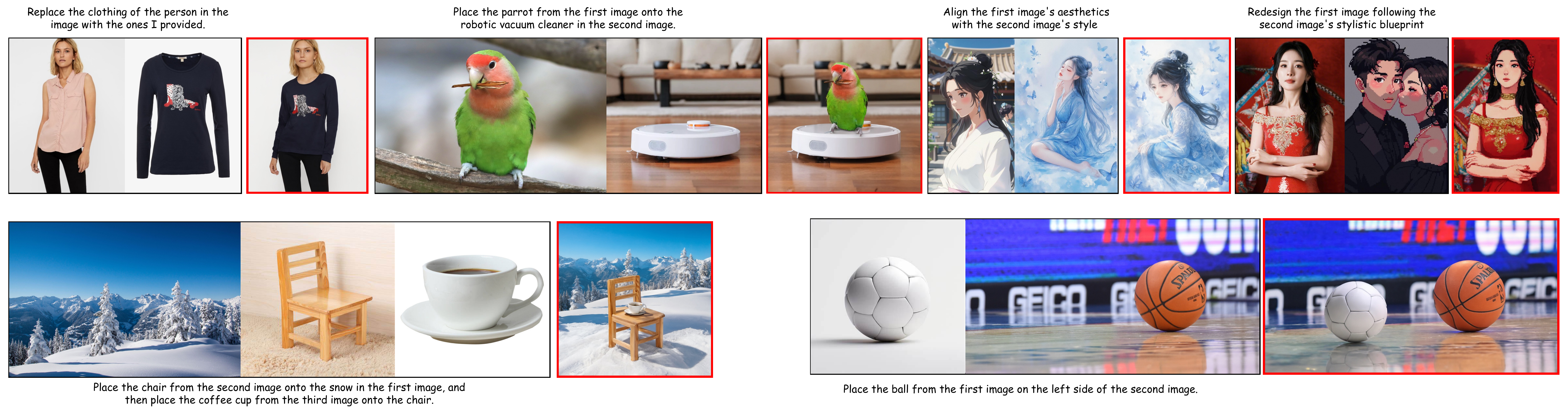}
    \caption{\textbf{More results on multi-element composition.} Our model can compose multiple reference subjects into a coherent scene while preserving the identity of each element.}
    \label{fig:multi_composition_app}
\end{figure*}

\begin{figure*}[t]
    \centering
    \includegraphics[width=\linewidth]{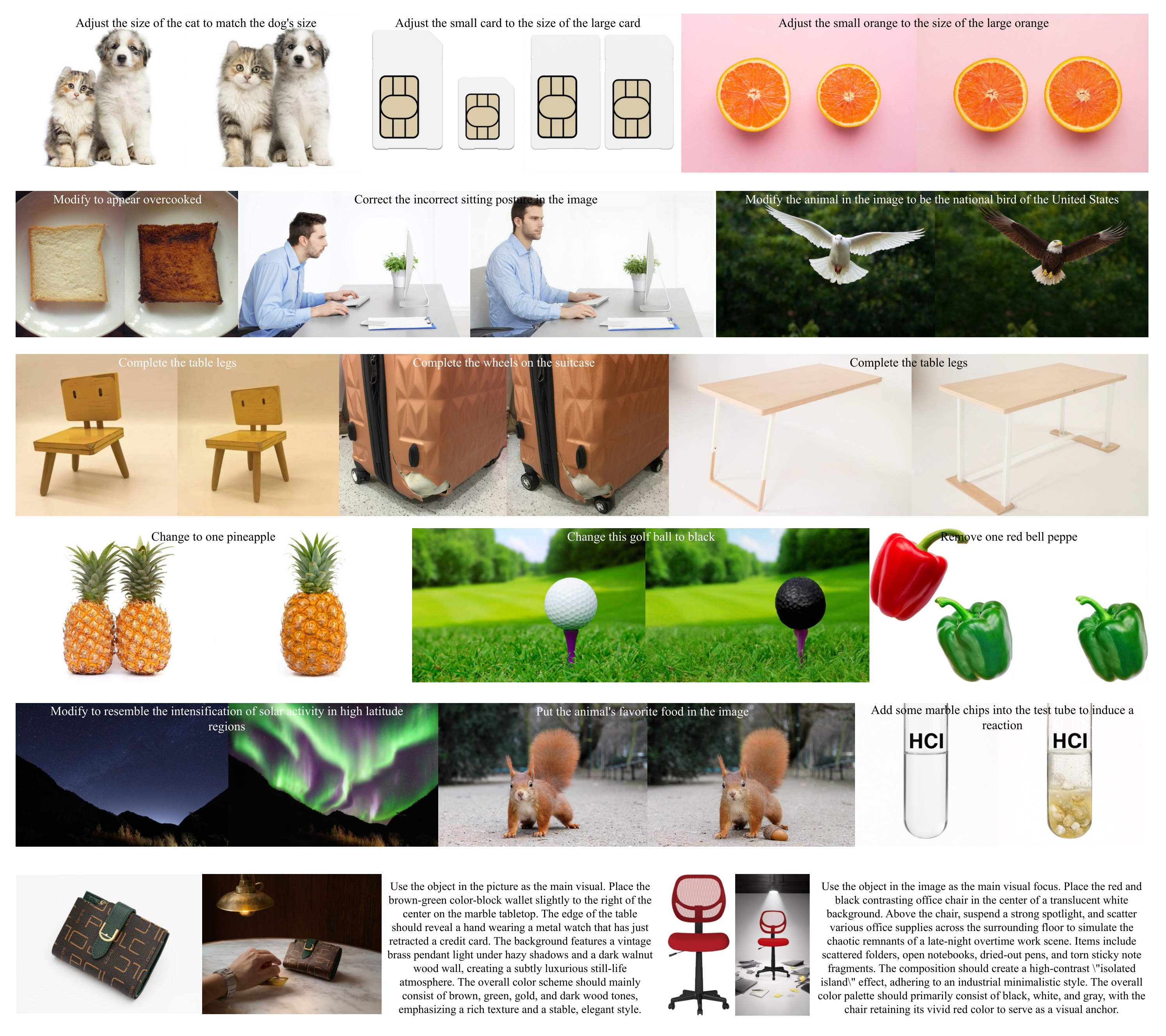}
    \caption{\textbf{More results on intelligent image editing.} These examples require complex world knowledge such as biological transformations, geographical understanding, and cultural context to perform accurate editing.}
    \label{fig:intelligent_editing_app}
\end{figure*}

We present comprehensive qualitative results to demonstrate the versatility of {\method} across diverse image editing and generation scenarios, spanning single-image editing (\Cref{fig:single_editing_app}), multi-element composition (\Cref{fig:multi_composition_app}), and knowledge-driven intelligent editing (\Cref{fig:intelligent_editing_app}).

\paragraph{Single Image Editing.}
\Cref{fig:single_editing_app} illustrates the broad editing capabilities of {\method} across six representative tasks: (1) \textit{Add/Remove/Extract}, which covers adding objects, removing elements, and extracting subjects from scenes; (2) \textit{Replace}, which substitutes objects or materials while faithfully preserving contextual layout; (3) \textit{Style Transfer}, which transforms image appearance into various artistic styles; (4) \textit{Background}, which modifies or replaces image backgrounds; (5) \textit{Subject-Driven Generation}, which generates new images conditioned on reference subjects; and (6) \textit{Controllable Generation}, which produces images guided by spatial or structural control signals. These results confirm that our compressed continuous semantic representation retains sufficient fine-grained information to enable precise, instruction-following edits.

\paragraph{Multi-Element Composition.}
\Cref{fig:multi_composition_app} demonstrates the ability of {\method} to compose multiple reference subjects into a coherent scene. Despite the challenge of maintaining the identity of each element while harmonizing spatial arrangement and stylistic consistency, our model produces visually plausible compositions. This capability benefits from the rich semantic priors preserved by the channel-compressed representation, enabling faithful identity preservation even without relying on VAE-based reconstruction.

\paragraph{Intelligent Image Editing.}
Beyond conventional instruction-based editing, we further evaluate {\method} on benchmarks that require world knowledge and contextual reasoning, namely WorldEdit~\cite{worldedit} and Kris~\cite{wu2025kris}. Unlike systems that rely on explicit editing commands, these benchmarks require models to integrate world knowledge priors and perform reasoning over semantic representations before generating the desired output. \Cref{fig:intelligent_editing_app} showcases representative examples that involve biological transformations, geographical understanding, and cultural context. Rather than following literal instructions, the model must infer what transformation is semantically appropriate given the input, which highlights the advantage of unifying understanding and generation within a shared semantic space.

Quantitatively, as shown in~\Cref{tab:mian}, UniCom achieves a high overall score of 4.35/5 on WorldEdit-Test, demonstrating strong performance across diverse cause categories. These results underscore the effectiveness of semantic representation for generation: by leveraging rich visual-language priors, {\method} produces more coherent and contextually accurate edits, outperforming most traditional methods that rely solely on explicit instructions. The integration of world knowledge and reasoning significantly enhances the model's ability to handle complex and context-sensitive editing tasks.

\begin{table}[ht]
\centering
\caption{The performance of both commercial and open-source models on image editing tasks in WorldEdit-Test, evaluated across different causes and metrics.}
\resizebox{\textwidth}{!}{
\begin{tabular}{c|c|ccc|ccccccccccc}
\toprule
\begin{tabular}[c]{@{}c@{}}Cause\\ Category\end{tabular} & Metric & GPT-4o  & Nano-Banana & Seedream4.0 & UniCom & WorldEdit & Flux-Kontext & Bagel-Think & Bagel  & Omnigen & Omnigen2 & Emu2 & Anyedit & Ip2p &Magicbrush \\
\midrule
\multirow{4}{*}{Time}                                         & VC                  &     4.02              &       4.22   & 3.92   &3.08  &   4.18      &     4.6      &  2.38   &   2.34      &   3.02   &   2.22       &     1.48              &    1.63         &  3.04 &   1.40         \\
                                                              & VQ                  &     5.00 &5.00          &       4.86  &4.78           &    4.55     &  4.88        & 4.52     &   4.02      &  4.20    &  4.35        &     4.56              &      3.25       &    4.51  &   4.28      \\
                                                              & IF                   &   3.86       &     3.90      & 4.25     &4.08        &    3.71     &  1.46         &  2.46    &  1.66       &  1.38    &    1.33      &      1.54             &        1.79     &   1.51      & 1.19     \\
                                                              & KP                   & 4.20       &   4.14       &  4.41       &4.38         & 3.94        &      1.42     &  2.86    &  1.80       &  1.48    &   1.25       &     1.75              &  1.92           &   1.65      &   1.34   \\
                                                              \cline{2-16}
                                                              & Avg                 &    {4.27}            &   {4.32}   &   4.36     &4.08      &   {4.10}      &       3.09    &  3.06    &   2.46      & 2.52     &  2.29        &    2.33               &    2.15         & 2.68        & 2.05     \\
\midrule
\multirow{4}{*}{Temperature}                                  & VC                   & 4.02      &    4.22        &     4.06      &3.40      &  3.52       &  3.86         &  2.08    &  1.94       &  2.54    & 2.26         &     1.38              &  1.45           &   2.62      &  1.81    \\
                                                              & VQ                   & 5.00      &   4.96        &  4.78    &4.34            &   4.36      &   4.54        &   4.06   &  4.04       &  4.00    &  4.46        &    4.65               &   2.92          &   4.56      &  4.28    \\
                                                              & IF                   & 4.54     &   4.20        &    4.50      &4.38         & 4.04        &     2.08      &  3.52    &  1.80       &  1.54    &     1.46     &     1.29              &          1.51   &   1.32     & 1.28      \\
                                                              & KP                   &    4.66 &       4.26         &    4.48     &4.20      &  3.90       &       1.88    & 3.76     &   1.90      &  1.38    &   1.37       &   1.38                &  1.45           &  1.44       &1.30      \\
                                                              \cline{2-16}
                                                              & Avg                  & {4.56}      &   {4.41}         &     4.46     &4.06       &  3.96       &       3.09    &  3.36    &   2.42      & 2.37     &  2.39        &   2.17                &         1.83    &  2.49       &  2.17    \\
\midrule
\multirow{4}{*}{Humidity}                                     & VC                   &  4.50    &    4.40        &  4.40         &3.66       &   3.82      &   4.40        &  2.68    &   2.32      &  2.28    &  1.94        &      1.76             &    1.62         &  2.86     & 1.40       \\
                                                              & VQ                   &   4.94    &   4.90       &    4.94     &4.62          &   4.58      &      4.94     & 4.42     & 4.32        &   4.14   &    4.50      &       4.56            &  3.52           & 4.62      &   4.28     \\
                                                              & IF                   &     3.92    &  3.52     &    4.74       &4.56         &  3.28       &       1.52    &  3.36    & 2.22        &  1.34    &  1.54        &    1.88               &         1.90   &   1.72     &  1.36     \\
                                                              & KP                   &   4.26     &     3.76      &    4.78      &4.22       &  3.44       &       1.60    &  3.36    &  2.16       & 1.44     &  1.64        &   2.20                &         2.06    & 1.90        & 1.40     \\
                                                              \cline{2-16}
                                                              & Avg                  &   {4.41}       &        {4.15}      &     4.72     &4.20  &  3.78       &       3.12    &  3.46    &  2.76       &   2.30   &    2.41      &   2.60                &  2.28           & 2.78       &   2.11    \\
\midrule
\multirow{4}{*}{Acidity}                                      & VC                   &    4.46     &   4.36      &    3.92       &3.42       &   3.94      &    4.74       &  2.74    &  2.60       &  3.14    &   1.88       &          1.34         &     1.38        & 2.74       &  1.60     \\
                                                              & VQ                   &    5.00  &    4.76         &      4.94       &4.75    &   4.76      &   4.88        & 4.28     &   4.40      &   4.28   &  4.68        &         4.70          &          3.32   & 4.40    &  4.26       \\
                                                              & IF                  &  3.82    &    3.58        &  4.38      &4.02          &     3.50    &     1.20      &   2.36   & 1.40        &  1.10    &      1.08    &    1.12               &  1.22           &  1.18    &  1.12       \\
                                                              & KP                   &      3.70 &    3.44        &  4.06   &3.91            &  3.68       &     1.20      &  2.34    &   1.40      &  1.06    &   1.08       &      1.14             &  1.26           &  1.14       & 1.04     \\
                                                              \cline{2-16}
                                                              & Avg                  &    {4.25}  &   {4.04}          &     4.33      &4.03      & 3.97        &       3.01    & 2.93     &   2.45      &  2.40    &   2.18       &     2.08              &        1.80     & 2.37      &   2.01     \\
\midrule
\multirow{4}{*}{Light}                                        & VC                   &   3.70     &    4.40       & 4.24       &3.34         &  3.82       &   4.64        &   2.84   &  2.36       & 2.44     &   1.88       &      1.45             &    1.43         &   2.66      &  1.54    \\
                                                              & VQ                   &   4.78    &    4.92       &    4.96      &4.51        & 4.68        &      4.84     &  4.46    &  4.26       &  4.26    &   4.48       &   4.71                &        2.98     &   4.50   &   4.20      \\
                                                              & IF                   &   4.40    &   2.88       &  4.44       &3.77         &  3.74       &    1.86       &  3.60    &  2.64       & 1.98     &    2.31      &  2.22                 &         2.11    &  1.62     &  1.41      \\
                                                              & KP                   &   4.64     &   3.50      &     4.58     &4.46         &  4.12       &      2.22     & 3.84     &   3.00      & 2.20     &   2.67       &   2.67                &         2.28    &  2.06       &   1.83   \\
                                                              \cline{2-16}
                                                              & Avg                  &  {4.38} 
                                                              &  3.93          &      4.56     &4.02         &     {4.09}    &        3.39   & 3.69     &  3.07       &  2.72    &     2.83     &     2.77              &         2.20    &  2.71      &  2.25     \\
\midrule
\multirow{4}{*}{Break}                                        & VC                   &  4.80     &    4.90        &     4.64      &4.14      & 4.74        & 4.88          & 4.00     &  2.98       &2.98      & 1.80         &     1.36              &    1.46         &   2.36     &  1.40     \\
                                                              & VQ                   &   4.92     &  4.78       &  4.94    &4.82             &   4.48      &      4.72     & 4.12     &  4.06       & 4.54     &    4.46      &    4.64               &  3.28           &  4.50     & 4.14       \\
                                                              & IF                   & 4.06      &  4.34        & 4.58        &4.66          &   4.56      &     2.36      &  3.10    &  2.18       & 1.40     &    1.10      &     1.36              & 1.62            & 1.08     & 1.22        \\
                                                              & KP                   &    3.84   &   4.14         &     4.38     &4.61       &  4.36       &        2.06   &  2.72    &   1.92      &  1.24    &     1.02     &    1.30               &       1.46      & 1.10        & 1.14     \\
                                                              \cline{2-16}
                                                              & Avg                  &   {4.41}      &     {4.54}      &    4.64      &4.56      &  {4.54}       &       3.51    & 3.49     &  2.79       & 2.54     &    2.10      &   2.17                &        2.00     & 2.26       & 1.98      \\
\midrule
\multirow{4}{*}{Inflate}                                       & VC                   &   4.53      &  4.88       &   4.63      &3.78         & 4.20        &   4.54        &  3.00    &  2.53       & 2.38     & 1.94         &      1.59             & 1.31            &  2.58      &  1.56     \\
                                                              & VQ                   &    4.94       &   4.98    &  4.65     &4.58           &  4.46       &     4.92      & 4.46     &  4.12       & 3.98     &    4.47      & 4.57                  &        3.20     & 4.70       &  4.15     \\
                                                              & IF                   &    4.22    &   3.56      &    3.82       &4.56        &  3.96       &      1.48     &  3.34    &    3.14     & 2.12     &    2.67      &   2.33                &  2.08           & 1.64     &   1.65      \\
                                                              & KP                   &   4.31    &   3.36       &  3.51        &3.88         & 3.70        &      1.46     & 3.06     &    2.94     &  1.78    &   2.37       &   2.37                &  1.74           &   1.5      &  1.60    \\
                                                              \cline{2-16}
                                                              & Avg                  &  {4.50}       &   {4.20}      &     4.15       &4.20      &   4.08      &   3.10        & 3.47     &   3.18      & 2.57     &    2.86      &   2.71                &        2.08     &  2.61     &   2.24     \\
\midrule
\multirow{4}{*}{Squeeze}                                      & VC                  &  4.44     &     4.80        &     4.56    &3.52       &   4.14      &  4.68         & 3.12     &  2.46       &   2.92   &   2.26       &       1.53            &   1.48          &   2.36     &  1.79     \\
                                                              & VQ                   &   4.94    &    4.94       & 4.76       &4.77          &  4.34       &    4.82       &  3.88    &   4.12      & 4.24     &   4.62       &     4.51              & 3.22            &  4.24      &  4.27     \\
                                                              & IF                   &   2.86             &  2.54    &     3.78    &2.59     &  3.38       &       1.22    & 2.56     &   1.90      &  1.42    &   1.36       &     1.64              &         1.48    &  1.24     &   1.46     \\
                                                              & KP                   &    2.74 &   2.48         &     3.82        &3.62      &  3.04       &        1.22   & 2.28     &  1.70       &   1.32   &     1.34     &      1.30             &  1.30           &  1.20       &   1.38   \\
                                                              \cline{2-16}
                                                              & Avg                  &   {3.75}       &   3.69    &    4.23       &3.63        &   {3.73}      &       2.99    &  2.96    &  2.55       &  2.48    &    2.39      &   2.25                &  1.87           &   2.26    &   2.22     \\
\midrule
\multirow{4}{*}{Twist}                                        & VC                   &  4.52      &      4.52      &  4.68       &4.24       & 4.3        &    4.52       &  2.56    &   2.22      &   2.62   &   2.02       &     1.63              &   1.43          &  1.84       &  1.66    \\
                                                              & VQ                   &   4.90   &    5.00        &  4.86      &4.52          &  4.32       &     4.98      & 4.74     &  4.24       &  4.02    &   4.45       &    4.55               &        3.31     &  4.12     &  4.25      \\
                                                              & IF                   &    3.86     &   3.72     &     4.22       &3.99       &  4.14       &        1.78   &  3.00    &   2.38      & 1.76     &   1.61       &   1.59                &         1.82    & 1.12    &   1.46       \\
                                                              & KP                   &    3.80    &   3.94      &     3.80       &3.77       &   3.88      &        1.78   &   2.80   &  2.28       &  1.62    &   1.65       &   1.61                &   1.71          &  1.10       &   1.46   \\
                                                              \cline{2-16}
                                                              & Avg                 &  {4.27}  &  {4.30}            &    4.39    &4.13         & 4.16        &       3.27    &  3.28    &  2.78       & 2.51     &     2.43     &    2.35               &  2.07           &   2.05     &  2.21     \\
\midrule
\multirow{4}{*}{Stretch}                                      & VC                 &   4.36     &    4.34        &  4.26       &3.80       &  3.98       & 4.20          &  3.22    &   2.38      &  2.36    &  2.07        &        1.90           &    1.54         &   2.10      &  1.51   \\
                                                              & VQ                   &    5.00    &    4.98      & 4.90          &4.72       &   4.54      &       4.72    &4.52      &  3.98       &  4.14    &   4.54       &     4.59              &        3.10     &   4.10    & 3.79       \\
                                                              & IF                   &   4.44    &   3.84       &     4.06        &3.96      &   3.96      &     2.00      & 3.44     &  2.60       & 2.18     &    2.33      &    2.55               &  2.06           &  1.08     &  1.04      \\
                                                              & KP                   &  4.52     &     3.76       & 4.10         &4.18       &   4.10      &     2.12      &   3.48   &   2.50      &  2.10    &   2.48       &      2.71             & 2.04            &  1.08       &    1.21  \\
                                                              \cline{2-16}
                                                              & Avg                  &   {4.58}   &  {4.23}          &           4.33       &4.19         &  4.15         &  3.26    &  3.67       & 2.87     &       2.70   &     2.85              &   2.94          &   2.19     & 2.09 &1.89     \\
\midrule
\multirow{4}{*}{Causal}                                         & VC                   &  4.26       &     4.46       &4.56      &4.18         &   4.22      &  3.96         & 3.20     & 2.68        & 2.20     &  3.54        &     2.10              &   2.08          &   2.50       &  1.86   \\
                                                              & VQ                   &    4.88    &     4.84      &  4.98        &4.76       &     4.84    &    4.76       & 4.16     &  4.66       &  3.84    &   4.88       &      4.71             &         3.66    &   4.28     &  4.12     \\
                                                              & IF                   &    4.74 &   4.58         &     4.78     &4.35         &  4.04       &       2.50    &  3.50    &  2.54       & 2.18     &    1.56      &  1.88                 &  2.24           & 1.74     &  1.80       \\
                                                              & KP                   &   4.68    &   4.56       &     4.84     &4.13         &  3.96       &       2.70    &   3.38   &  2.52       &   2.10   &    1.56      &    1.90               &  2.20           &   1.72      &   1.96   \\
                                                              \cline{2-16}
                                                              & Avg                  &  {4.64}      &     {4.61}       &  4.79      &4.36        &  4.27       &     3.48      &  3.56    & 3.10        & 2.58     &   2.89       &    2.65               &      2.55   & 2.56   &  2.44   \\
\midrule
\rowcolor{mygreen}
Overall & Avg                  &    \underline{4.36}             &  {4.22}    & \textbf{4.45}     &{4.12}    & 4.07      &  3.21      &   3.35   &  2.76     & 2.52  &     2.51     &    2.46              & 2.09     &  2.44  &2.14  \\
\bottomrule
\end{tabular}
}
\label{tab:mian}
\end{table}

\section{Chain-of-Thought Enhanced Generation}
\label{app:cot}

To further improve the quality and controllability of image editing and generation, we incorporate chain-of-thought (CoT) reasoning into the training pipeline of {\method}. Specifically, during training, we enable CoT for a proportion of samples by prepending a \texttt{<think>} token before the output image. This encourages the model to first produce a structured reasoning trace that plans the visual transformation, and then generate the target image conditioned on both the original input and the reasoning content.

\begin{figure*}[t]
    \centering
    \includegraphics[width=\linewidth]{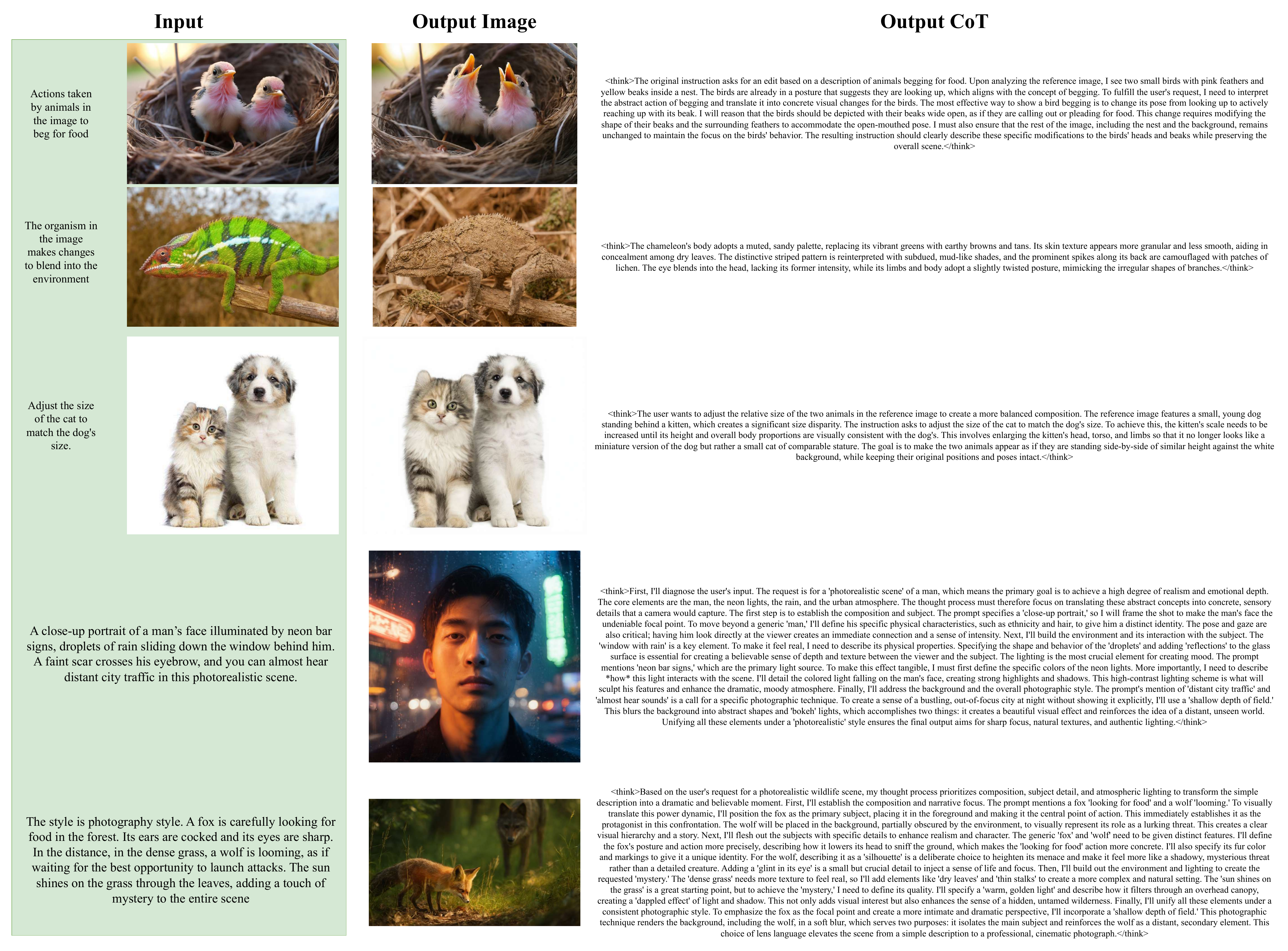}
    \caption{\textbf{Chain-of-thought enhanced generation and editing.} For each example, we show the generated image along with the corresponding thinking process enclosed in \texttt{<think>} tokens. The model first reasons about the required visual transformation, then generates the output image accordingly.}
    \label{fig:cot_samples}
\end{figure*}

\Cref{fig:cot_samples} presents representative examples where CoT is activated during inference. For each case, we display the generated image alongside the corresponding thinking process. As shown, the model explicitly reasons about the required modifications (e.g., identifying which attributes to change, understanding spatial relationships, or decomposing complex instructions into sub-steps) before producing the output. This explicit reasoning step enables more accurate and semantically faithful generation, particularly for tasks that involve multi-step transformations or require fine-grained control over the output.

\section{Limitation}
While UniCom represents a significant step forward in unifying visual understanding and generation, there are still several limitations that need to be addressed.  One primary limitation is the complexity involved in training high-dimensional continuous representations.  Despite the success of our compression techniques, there is still potential for information loss during the compression process, particularly when dealing with very fine-grained details in complex images.  While we have demonstrated the ability of UniCom to preserve these details, the trade-off between compression and fidelity remains a challenging aspect of our approach.

Another limitation arises from the inherent challenges of training generative models, UniCom still requires substantial computational resources, especially for large-scale datasets and high-resolution image tasks.  The energy consumption associated with training such large models may also limit the accessibility and sustainability of this approach in resource-constrained environments.  Future work will need to address these scalability concerns, possibly through model distillation or optimization techniques.


\end{document}